\documentclass[10pt,journal,letterpaper,compsoc]{IEEEtran}
\usepackage{amsmath,amsfonts,amssymb,epsfig,verbatim,color}
\usepackage{graphicx}
\usepackage{multirow}
\usepackage{rotating}
\usepackage{booktabs}
\usepackage{pdfpages}
\usepackage{amsthm} 
\usepackage{subfig}
\usepackage{algpseudocode}
\usepackage{url}
\usepackage{bm}
\usepackage{array}
\usepackage{xcolor,colortbl}
\usepackage{textcomp}
\usepackage[ruled,norelsize]{algorithm2e}

\newcommand{\mat}[1]{\mathtt #1}

\newcommand{\vct}[1]{\mathbf #1}
\newcommand{\T}{\ensuremath{^\top}}
\newcommand{\argmax}{\operatornamewithlimits{\arg\,\max}}

\newcommand{\x}{{\bf x}}
\newcommand{\y}{{\bf y}}

\newcommand{\etal}{\textit{et al. }}
\newtheorem{proposition}{proposition}
\newtheorem{Definition}{Definition}

\makeatletter
\newcommand{\removelatexerror}{\let\@latex@error\@gobble}
\makeatother

\DeclareMathAlphabet{\mathpzc}{OT1}{pzc}{m}{it}

\DeclareMathAlphabet{\mathpzc}{OT1}{pzc}{m}{it}

\begin{document}
	
	\title{ Multi-Target Tracking in Multiple Non-Overlapping Cameras using Constrained Dominant Sets }
	
	\author{ Yonatan~Tariku~Tesfaye*,~\IEEEmembership{Student Member,~IEEE,}
		Eyasu~Zemene*,~\IEEEmembership{Student Member,~IEEE,}
		Andrea~Prati,~\IEEEmembership{Senior member,~IEEE,}
		Marcello~Pelillo,~\IEEEmembership{Fellow,~IEEE,}
		and~Mubarak~Shah,~\IEEEmembership{Fellow,~IEEE}
		\IEEEcompsocitemizethanks{
			
			\IEEEcompsocthanksitem {* The first and second authors have equal contribution.}
			\IEEEcompsocthanksitem Y. T. Tesfaye is with the department of Design and Planning in Complex Environments of the University IUAV of Venice, Italy. E-mail: y.tesfaye@stud.iuav.it
			
			\IEEEcompsocthanksitem E. Zemene and M. Pelillo are with the department of Computer Science, Ca' Foscari University of Venice, Italy. E-mail: \{eyasu.zemene, pelillo\}@unive.it
			
			\IEEEcompsocthanksitem A. Prati is with the department of Department of Engineering and Architecture of the University of Parma, Italy. E-mail: andrea.prati@unipr.it

			\IEEEcompsocthanksitem M. Shah is with the Center for Research in Computer Vision (CRCV), University of Central Florida, USA. E-mail: \{haroon,shah\}@eecs.ucf.edu}}

	\markboth{}%
	{Shell \MakeLowercase{\textit{et al.}}: Bare Demo of IEEEtran.cls for Computer Society Journals}
	
	\IEEEcompsoctitleabstractindextext{

		\begin{abstract}
			
			In this paper, a unified three-layer hierarchical approach for solving tracking problems in multiple non-overlapping cameras is proposed. Given a video and a set of detections (obtained by any person detector), we first solve {\em within-camera tracking} employing the first two layers of our framework  and, then, in the  third layer, we solve {\em across-camera tracking} by  merging tracks  of the same person in all cameras in a simultaneous fashion. To best serve our purpose, a constrained dominant sets clustering (CDSC) technique, a parametrized version of standard quadratic optimization, is employed to solve both tracking tasks. The tracking problem is caste as finding constrained dominant sets from a graph. That is, given a constraint set and a graph, CDSC  generates cluster (or clique), which forms a compact and coherent set that contains a subset of the constraint set. The approach is based on a parametrized family of quadratic programs that generalizes the standard quadratic optimization problem. In addition to having a unified framework that simultaneously solves within- and across-camera tracking, the third layer helps link broken tracks of the same person occurring during within-camera tracking.  A standard algorithm to extract constrained dominant set from a graph is given by the so-called replicator dynamics whose computational complexity is quadratic per step which makes it handicapped for large-scale applications. In this work, we  propose a fast algorithm, based on dynamics from evolutionary game theory,  which is efficient and salable to large-scale real-world applications. We have tested this approach on a very large and challenging dataset (namely, MOTchallenge DukeMTMC) and show that the proposed framework outperforms the current state of the art. Even though the main focus of this paper is on multi-target tracking in non-overlapping cameras, proposed approach can also be applied to solve {\em re-identification} problem. Towards that end, we also have performed  experiments on MARS, one of the largest and challenging video-based person re-identification dataset, and have obtained excellent results. 
			These experiments demonstrate the general applicability of the proposed framework for non-overlapping across-camera tracking and  person re-identification tasks.
			
		\end{abstract}
		
		}

	\maketitle

	\section{Introduction}
	As the need for visual surveillance grow, a large number of cameras have been deployed to cover large and wide areas like airports, shopping malls, city blocks etc.. Since the fields of view of single cameras are limited,  in most wide area surveillance scenarios, multiple cameras are required to cover larger areas. Using multiple cameras with overlapping fields of view is costly from both economical and computational aspects. Therefore, camera networks with non-overlapping fields of view are preferred and widely adopted in real world applications.
	
	In the work presented in this paper, the goal is to track  multiple targets  and maintain their identities as they move from one camera to the another camera with non-overlapping fields of views. In this context, two problems need to be solved, that is, within-camera data association (or tracking)  and across-cameras data association by employing the tracks obtained from within-camera tracking. Although there have been significant progresses in both problems separately, tracking multiple target jointly in both within and across non-overlapping cameras remains a less explored topic.  Most approaches, which solve multi-target tracking in multiple non-overlapping cameras \cite{YinKaiTieICPR08,AmiPanSurWeiBMVC04, OmaZeeKhuMubICCV03, YouSenMusMTA14, CheBhaTCS16}, assume tracking within each camera has already been performed and try to solve tracking problem only in  non-overlapping cameras; the results obtained from  such approaches are far from been optimal \cite{YouSenMusMTA14}. 
	
	\begin{figure*}
		\centering 
		\includegraphics[width=1\linewidth,trim=0cm 2.88cm 0cm 2.88cm,clip]{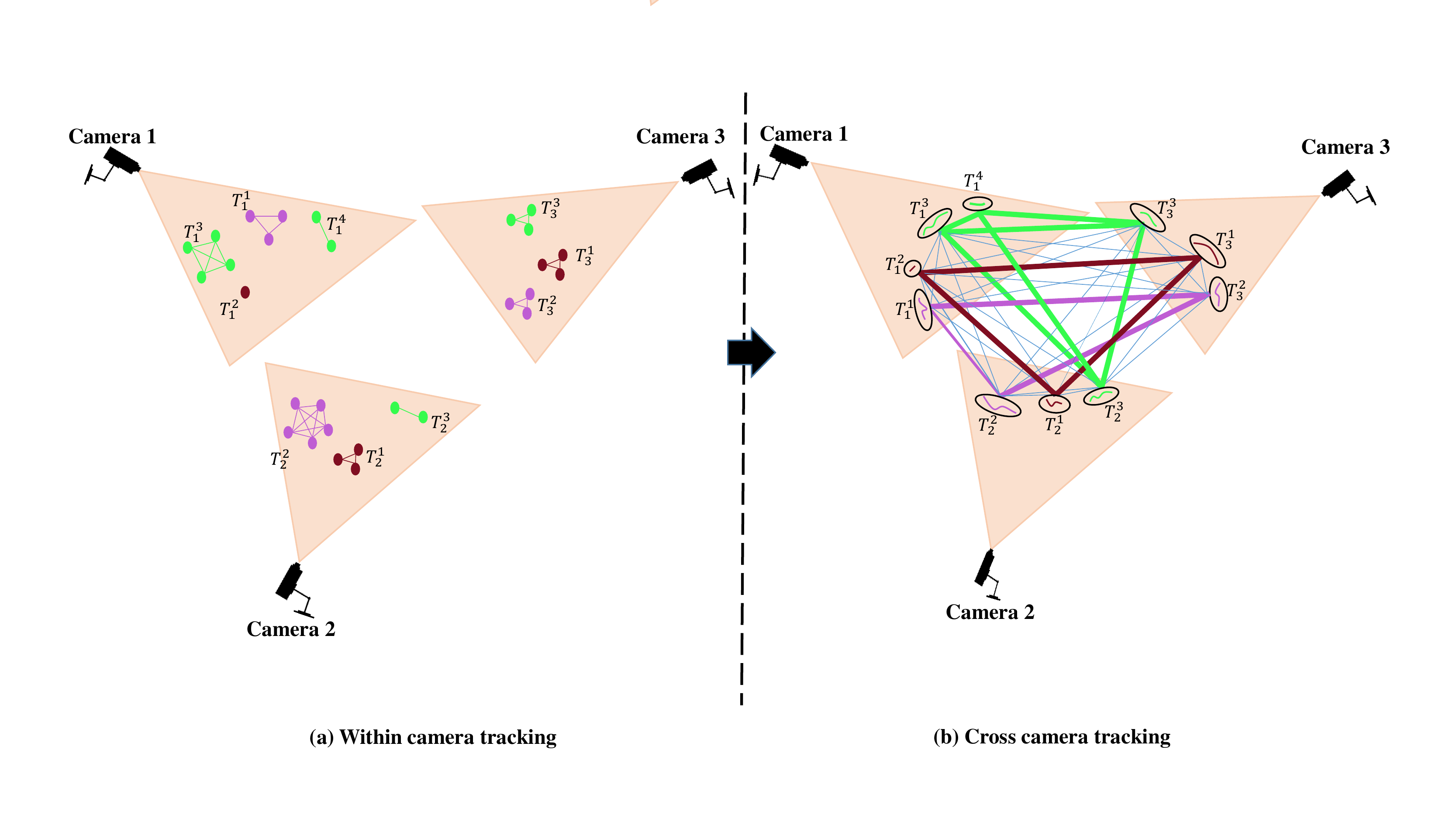}
		\caption{A general idea of the proposed framework. (a) First, tracks are determined within each camera, then (b) tracks of the same person from different non-overlapping cameras are associated, solving the across-camera  tracking. Nodes in (a) represent tracklets and nodes in (b) represent tracks. The $i^{th}$ track of camera $j$, $T^i_j$, is a set of tracklets that form a clique. In (b) each clique in different colors represent tracks of the same person in non-overlapping cameras. Similar color represents the same person. (Best viewed in color)}
		\label{fig:overview}
	\end{figure*}
	
	In this paper, we propose a hierarchical approach in which we first determine tracks within each camera, (Figure \ref{fig:overview}(a)) by solving data association, and later we associate  tracks of the same person in different cameras in a unified approach (Figure \ref{fig:overview}(b)), hence solving the across-camera tracking. Since appearance and motion cues of a target tend to be consistent in a short temporal window in a single camera tracking,  solving tracking problem in a hierarchical manner is common: tracklets are generated within short temporal window first and later they are merged  to form  full tracks (or trajectories) \cite{AmiShaECCV12, gmmcp, YonEyaPelPraIET2016}. Often, across-camera tracking  is more challenging than solving within-camera tracking due to the fact that appearance of people may exhibit significant differences due to illumination variations and pose changes between cameras.

	Therefore, this paper proposes a unified three-layer framework to solve both within- and across-camera tracking. In the first two layers, we generate tracks within each camera and in the third layer we  associate all tracks of the same person across all cameras in a simultaneous fashion.

	To best serve our purpose, a constrained dominant sets clustering (CDSC) technique, a parametrized version of standard quadratic optimization, is employed to solve both tracking tasks. The tracking problem is caste as finding constrained dominant sets from a graph. That is, given a constraint set and a graph, CDSC  generates cluster (or clique), which forms a compact and coherent set that contains all or part of the constraint set. \textit{Clusters} represent tracklets and tracks in the first and second layers, respectively. The proposed within-camera tracker can robustly handle long-term occlusions,  does not change the scale of original problem as it does not remove nodes from the graph during the extraction  of compact clusters and is several orders of magnitude faster (close to real time) than existing methods. Also, the proposed across-camera tracking method using CDSC and later followed by refinement step offers several advantages. More specifically, CDSC not only considers the affinity (relationship) between tracks, observed in different cameras, but also takes into account the affinity  among tracks from the same camera. As a consequence, the proposed approach not only accurately associates  tracks from different cameras but also makes it possible to link multiple short broken tracks obtained during within-camera tracking, which may belong to a single target track.  
	For instance, in Figure \ref{fig:overview}(a)  track $T_1^3$ (third track from camera 1) and $T_1^4$ (fourth track from camera 1) are tracks of same person which were mistakenly broken from a single track. However, during the third layer, as they are highly similar to tracks in camera 2 ($T_2^3$) and camera 3 ($T_3^3$), they form a clique, as shown in Figure \ref{fig:overview}(b).   Such across-camera formulation is able to associate these broken tracks with the rest of tracks from different cameras, represented with the green cluster in Figure \ref{fig:overview}(b).

	The contributions of this paper are summarized as follows:
	\begin{itemize}
		\item We formulate multi-target tracking in  multiple non-overlapping cameras as finding constrained dominant sets from a graph. We propose a three-layer hierarchical approach, in which we first solve within-camera tracking using the first two layers, and using the third layer we solve the across-camera tracking problem.  
		\item We propose a technique to further speed up our optimization by reducing the search space, that is, instead of running the dynamics over the whole graph, we localize it on the sub graph selected using the dominant distribution, which is much smaller than the original graph. 
		\item Experiments are performed on  MOTchallenge DukeMTMCT dataset and  MARS dataset,  and show improved effectiveness of our method with respect to the state of the art.
		
	\end{itemize}
	
	The rest of the paper is organized as follows. In Section \ref{Related Work}, we review relevant previous works. Overall proposed approach for within- and across-cameras tracking modules is summarized in section \ref{approach}, while sections \ref{within-camera tracking} and  \ref{across-camera tracking} provide more in details of the two modules. In section \ref{speedin up data association }, we present the proposed approach to further speed up our method. Experimental results are presented in Section \ref{experiments}. Finally, section \ref{conclusion}  concludes the paper.
	
	\section{Related Work}\label{Related Work}
	
	Object tracking is a challenging computer vision problem and has been one of the most active research areas for many years. In general, it can be divided in two broad categories: tracking in single and  multiple cameras. Single camera object tracking associates object detections across frames in a video sequence, so as to generate the object motion trajectory over time. Multi-camera tracking aims to solve handover problem from one camera view to another and hence establishes target correspondences among  different cameras, so as to achieve consistent object labelling across all the camera views. Early multi-camera target tracking research works fall in different categories as follows. Target tracking with partially overlapping camera views has been researched extensively during the last decade \cite{NadAndAVSS09, calderara2008bayesian, CarRauNarLuiFerACM08, SohMubPAMI03, BirThoGerICPR08, SenJasCheWayJasIVP08}. Multi target tracking across multiple cameras with disjoint views has also been researched in \cite{YinKaiTieICPR08,AmiPanSurWeiBMVC04, OmaZeeKhuMubICCV03, YouSenMusMTA14, CheBhaTCS16}. Approaches for overlapping field of views  compute spatial proximity of tracks in the overlapping area, while approaches for tracking targets across cameras with disjoint fields of view, leverage appearance cues together with spatio-temporal information.

	Almost all  early multi-camera research works try to address only across-camera tracking problems, assuming that within-camera tracking results for all cameras are given. Given tracks from each camera, similarity among tracks is computed and target correspondence across cameras is solved, using the assumption that a track of a target in one camera view can match with at most one target track in another camera view.  Hungarian algorithm \cite{KuhHarNRLQ56} and  bipartite graph matching \cite{OmaZeeKhuMubICCV03} formulations are  usually used to solve this problem. Very recently, however, researchers have argued that assumptions of cameras having overlapping fields of view and the availability of intra-camera tracks  are unrealistic \cite{YouSenMusMTA14}. Therefore, the work proposed in this paper addresses the more realistic problem by solving both within- and across-camera tracking in one joint framework. 
	
	In the rest of this section, we first review the most recent works for single camera tracking, and then describe the previous related works on multi-camera multi-view tracking.

	Single camera target tracking associates target detections across frames in a video sequence in order to generate the target motion trajectory over time.
	Zamir \etal \cite{AmiShaECCV12} formulate tracking problem as generalized maximum clique problem (GMCP), where the relationships between all detections in a temporal window are considered. In \cite{AmiShaECCV12}, a cost to each clique is assigned and  the selected clique maximizes a score function. Nonetheless, the approach  is prone to local optima as it uses greedy local neighbourhood search. Deghan \etal \cite{gmmcp} cast tracking as a generalized maximum multi clique problem  (GMMCP) and follow a joint optimization for all the tracks simultaneously. To handle outliers and weak-detections associations they introduce  dummy nodes. However, this solution is computationally expensive. In addition, the hard constraint in their optimization makes the approach impractical for large graphs. Tesfaye \etal \cite{YonEyaPelPraIET2016} consider all the pairwise relationships between detection responses in a temporal sliding window, which is used as an input to their optimization based on fully-connected edge-weighted graph. They formulate tracking as finding dominant set clusters. Though the dominant set framework is effective in extracting compact sets from a graph \cite{EyaYonHarAndMarMubPAMI}\cite{PavPelPAMI07}\cite{EyaYonAndPeliICPR2016} \cite{BulPelBomCVIU11} \cite{MeqBulPelSIMBAD2015}, it follows a pill-off strategy to enumerate all possible clusters, that is, at each iteration it removes the found cluster from the graph which results in a change in scale (number of nodes in a graph) of the original problem.  
	In this paper, we propose a multiple target tracking approach, which in contrast to  previous works, does not need additional nodes to handle occlusion nor encounters change in the scale of the problem.

	Across-camera tracking aims to establish target correspondences among trajectories from different cameras so as to achieve consistent target labelling across all camera views. It is a challenging problem due to the illumination and pose changes across cameras, or track discontinuities due to the blind areas or miss detections. Existing  across-camera tracking methods try to deal with the above problems  using appearance cues. The variation in illumination of the appearance cues has been leveraged using different techniques such as Brightness Transfer Functions (BTFs). To handle the appearance change of a target as it moves from one camera to another, the authors in  \cite{OmaKhuZeeMubCVPR08} show that all brightness transfer functions from a given camera to another camera lie in a low dimensional subspace, which is learned by employing probabilistic principal component analysis and used for appearance matching. Authors of \cite{AndRicECCV06} used an incremental
	learning method to model the colour variations and \cite{BryShaTaoBMVC08} proposed a Cumulative Brightness Transfer Function, which is a better use of the available colour information from a very sparse training set. Performance comparison of different variations of Brightness Transfer Functions can be found in \cite{TizPiePaoICDSC09}. Authors in \cite{SatKaEdwTCASSP11} tried to achieve color consistency using colorimetric principles, where the image analysis system is modelled as an observer and  camera-specific transformations are determined, so that images of the same target appear similar to this observer. Obviously,  learning Brightness Transfer Functions or color correction models requires large amount of training data and they may not be robust against drastic illumination changes across different cameras. Therefore,  recent approaches have combined them with spatio-temporal cue which improve multi-target tracking performance \cite{DeYihJinQiqNanNeuro17, CheChaRamECCV10,YueRonLonAle,ShuYinAmiCVIU15,YinGerWACV14,XiaKaiTiePR14}. Chen \etal \cite{DeYihJinQiqNanNeuro17} utilized human part configurations for every target track from different cameras to describe the across-camera spatio-temporal constraints for across-camera track association, which is formulated as a multi-class classification problem via Markov Random Fields (MRF). Kuo \etal \cite{CheChaRamECCV10} used Multiple Instance Learning (MIL) to learn an appearance model, which effectively combines multiple image descriptors and their corresponding similarity measurements. The proposed appearance model combined with spatio-temporal information improved across-camera track association solving the “target handover” problem across cameras. Gao \etal \cite{YueRonLonAle} employ tracking results of different trackers and use their spatio-temporal correlation, which help them enforce tracking consistency and establish pairwise correlation among multiple tracking results. Zha \etal \cite{ShuYinAmiCVIU15} formulated tracking of multiple interacting targets as a network flow problem, for which the solution can be obtained by the K-shortest paths algorithm. Spatio-temporal relationships among targets are utilized to identify group merge and split events. In \cite{YinGerWACV14} spatio-temporal context is used  for collecting samples for discriminative appearance learning, where target-specific appearance models are learned to distinguish different people from each other. And the relative appearance context models inter-object appearance similarities for people walking in proximity and helps disambiguate individual appearance matching across cameras. 
	
	The problem of target tracking across multiple non-overlapping cameras is also tackled in \cite{RisSolZouCucTomECCV16} by extending their previous single camera tracking method \cite{ErgCarACCV2014}, where they formulate the tracking task as a graph partitioning problem. Authors in \cite{XiaKaiTiePR14}, learn across-camera transfer models including both spatio-temporal and appearance cues. While a color transfer method is used to model the changes of color across cameras for learning across-camera appearance transfer models, the spatio-temporal model is learned using an unsupervised topology recovering approach.  Recently Chen \etal \cite{CheBhaTCS16} argued that low-level information (appearance model and spatio-temporal information) is unreliable for tracking across non-overlapping cameras, and integrated contextual information such as social grouping behaviour. They formulate tracking using an online-learned Conditional Random Field (CRF), which favours track associations that maintain group consistency. In this paper, for tracks to be associated, besides their high pairwise similarity (computed using appearance and spatio-temporal cues), their corresponding constrained dominant sets should also be similar. 

	Another recent popular research topic, video-based person re-identification(ReID)  \cite{JinAncXianWeiCVPR16, NiaJesPaulCVPR16, TaiShaXiaECCV14, DunCatLouLouICIAP09,LiaHuZhuLiCVPR15,FarBazPerMurCriCVPR10,XioGouCamSznECCV14,ZheSheTiaWanWanTiaICCV15,MaSuJurIVC14}, is closely related to across-camera multi-target tracking. Both problems aim to match tracks of the same persons across non-overlapping cameras. However, across-camera tracking aims at 1-1 correspondence association  between tracks of different cameras. Compared to most video-based ReID approaches, in which only  pairwise similarity between the probes and gallery is exploited, our across-camera tracking framework not only considers the relationship between probes and gallery but it also takes in to account the relationship among tracks in the gallery.
	
	\section{Overall Approach} \label{approach}

	In this section,  first we briefly introduce the basic definitions and properties of constrained dominant set clustering. This is followed by formulation of within- and across-camera tracking.

	\subsection{Constrained Dominant Set clustering. } 
	As introduced in \cite{ZemPelECCV16}, constrained dominant set clustering, a constrained quadratic optimization program, is an efficient and accurate approach, which has been applied for interactive image segmentation. The approach generalizes dominant set framework \cite{PavPelPAMI07}, which is a well known generalization of the maximal clique problem to edge weighted graphs. Given an edge weighted graph $G(V,E,w)$ and a constraint set $\mathcal{Q} \subseteq V$, where $V, E$ and $w$, respectively, denote the set of nodes (of cardinality $n$), edges and edge weights. The objective is to find the sub-graph that contains all or some of elements of the constraint set, which forms a coherent and compact set. 
	
	In our formulation, in the first layer, each node in our graph represents a  short-tracklet along a temporal window (typically 15 frames). Applying constrained dominant set clustering here aim at determining cliques in this graph, which correspond to  tracklets. Likewise, each node in a graph in the second layer represents a tracklet, obtained from the first layer, and CDSC is applied here to determine cliques, which correspond to tracks. Finally, in the third layer, nodes in a graph correspond to tracks from different non-overlapping cameras, obtained from the second layer, and CDSC is applied to determine cliques, which relate  tracks of the same person across non-overlapping cameras.
	Consider a graph, $G$, with $n$ vertices (set $ V $), and its weighted adjacency matrix $\mat{A}$. Given a parameter $ \alpha > 0$, let us define the following parametrized quadratic program:
	\begin{equation}
		\label{eqn:parQP}
		\begin{array}{ll}
			\text{maximize }  &  f_{\mathcal{Q}}^\alpha(\x) = \x\T ({\mat{A}} - \alpha I_{\mathcal{Q}}) \x, \\
			\text{subject to} &  \mathbf{x} \in \Delta,
		\end{array}
	\end{equation}
	\noindent where $\Delta=\{ \x \in \mathbb{R}^n: \sum_i x_i = 1, \text{ and } x_i \geq 0 \text{ for all } i=1 \ldots n\}$, $\mathbf{x}$ contains a membership score for each node and  $I_{\mathcal{Q}}$ is the $n \times n$ diagonal matrix whose diagonal elements are set to 1 in correspondence to the vertices contained in $ V \setminus \mathcal{Q}$ (a set $V$ without the element $\mathcal{Q}$) and to zero otherwise.
	
	Let $\mathcal{Q} \subseteq  V$, with $\mathcal{Q} \neq \emptyset$ and let $\alpha > \lambda_{\max}( {\mat{A}}_{ V \setminus \mathcal{Q}}) $, where $\lambda_{\max}( {\mat{A}}_{ V \setminus \mathcal{Q}})$ is the largest eigenvalue of the principal submatrix of ${\mat{A}}$ indexed by the elements of $ V \setminus \mathcal{Q}$.
	If $\x$ is a local maximizer of $f_{\mathcal{Q}}^\alpha$ in $\Delta$, then
	$\sigma(\x) \cap \mathcal{Q} \neq \emptyset$, where,
	$
	\sigma(\x) = \{i \in V~:~x_i > 0\}
	$ .
	
	The above result provides us with a simple technique to determine dominant set clusters containing user-specified query vertices, $\mathcal{Q}$. Indeed, if $\mathcal{Q}$ is a vertex selected by the user, by setting
	\begin{equation}
		\label{alphabound}
		\alpha > \lambda_{\max}({\mat{A}}_{ V \setminus \mathcal{Q}}),
	\end{equation}
	we are guaranteed that all local solutions of (\ref{eqn:parQP}) will have a support
	that necessarily contains elements of $\mathcal{Q}$.

	\subsection{Within-Camera Tracking} \label{within-camera tracking}
\begin{figure}[!h]
		\centering 
		\includegraphics[width=9cm,height=9cm]{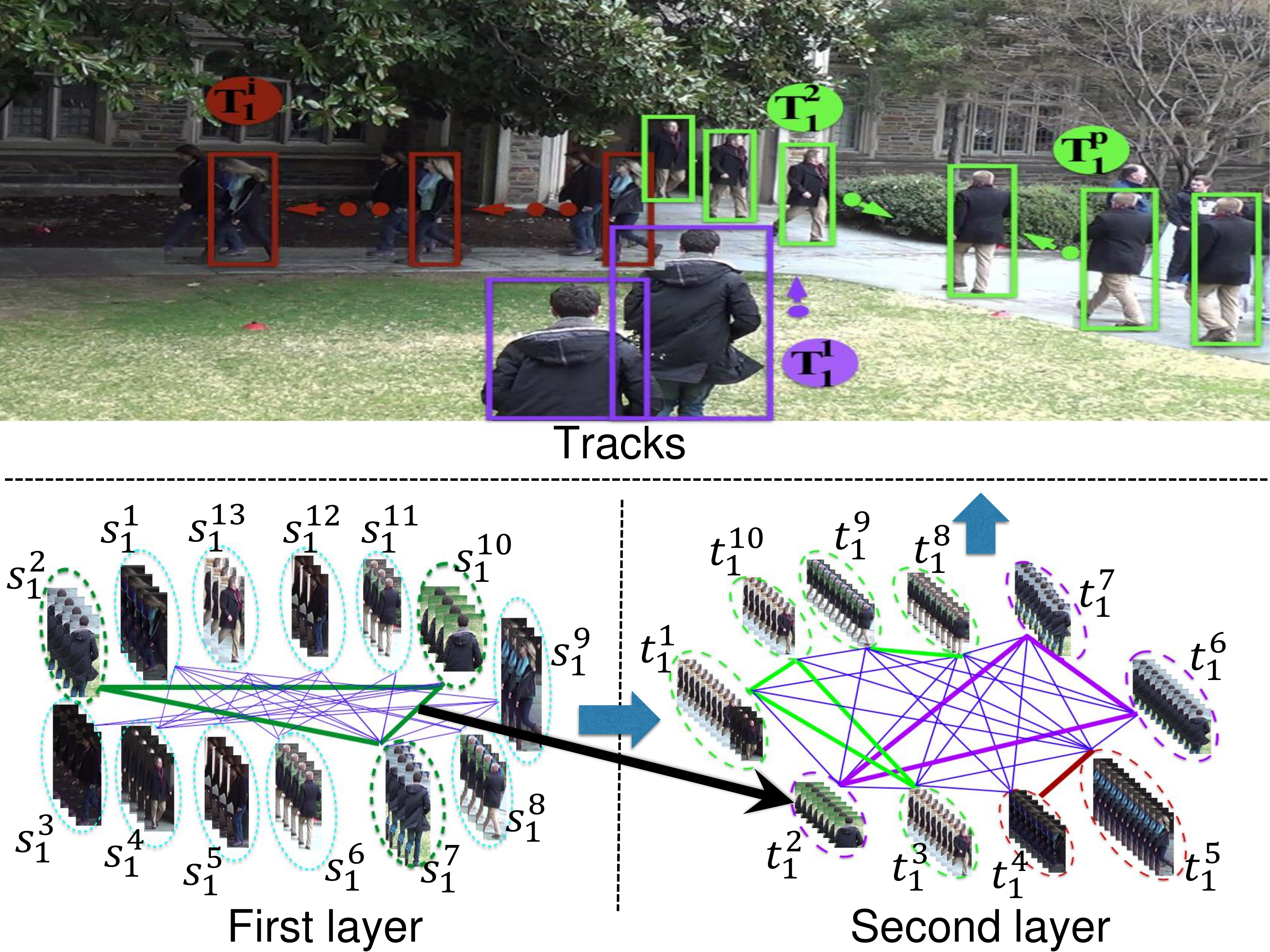}
		\caption{The figure shows within-camera tracking where short-tracklets from different  segments are used as input to our first layer of tracking. The resulting tracklets from the first layer are inputs to the second layer, which determine a tracks for each person. The three dark green short-tracklets ($s_1^2, s_1^{10}, s_1^7$), shown by dotted ellipse in the first layer, form a cluster resulting in tracklet ($t_1^2$) in the second layer, as shown with the black arrow. In the second layer, each cluster, shown in purple, green and dark red colors, form tracks of different targets, as can be seen on the top row. tracklets and tracks with the same color indicate same target. The two green cliques (with two tracklets and three tracklets) represent tracks of the person going in and out of the building (tracks $T^p_1$ and $T^2_1$ respectively) }
		
		\label{fig:within_camera_2}
	\end{figure}

	Figure \ref{fig:within_camera_2} shows proposed within-camera tracking framework. First, we divide a video into multiple short segments, each segment contains 15 frames, and generate short-tracklets, where human detection bounding boxes in two consecutive frames with 70\%  overlap, are connected \cite{gmmcp}. Then, short-tracklets from 10 different non-overlapping segments are used as input to our first layer of tracking. Here the nodes are short-tracklets (Figure \ref{fig:within_camera_2}, bottom left). Resulting tracklets from the first layer are used as an input to the second layer, that is, a tracklet from the first layer is now represented by a node in the second layer (Figure \ref{fig:within_camera_2}, bottom right). In the second layer, tracklets of the same person from different segment are associated forming tracks of a person within a camera. 
	
	\subsubsection{Formulation Using Constrained Dominant Sets}
	
	We build an input graph, $G(V,E,w)$,  where nodes represent short-tracklet ($s_i^j$, that is, $j^{th}$ short-tracklet of camera $i$) in the case of first layer (Figure \ref{fig:within_camera_2}, bottom left) and tracklet ($t_k^l$, that is, $l^{th}$ tracklet of camera $k$), in the second layer (Figure \ref{fig:within_camera_2}, bottom right). The corresponding affinity matrix  $\mat{A}=\left\{a_{i,j}\right\}$,  where $a_{i,j}=w(i,j)$ is built. The weight $w(i,j)$ is assigned to each edge, by considering both motion and appearance similarity between the two nodes. Fine-tuned CNN features are used to model the appearance of a node.  These features are extracted from the last fully-connected layer of Imagenet pre-trained 50-layers Residual Network (ResNet 50) \cite{KaiXiaShaJiaCVPR2016} fine-tuned using the trainval sequence of DukeMTMC dataset. Similar to \cite{AmiShaECCV12}, we employ a global constant velocity model to compute motion similarity between two nodes.

	\textbf{Determining cliques}: In our formulation, a clique of graph $G$  represents  tracklet(track) in the first  (second) layer. 
	Using short-tracklets/tracklets as a constraint set (in eq. \ref{eqn:parQP}), we enumerate all clusters, using game dynamics, by utilizing intrinsic properties of constrained dominant sets. Note that we do not use peel-off strategy to remove the nodes of found cliques from the graph, this keeps the scale of our problem (number of nodes in a graph) which guarantees that all the found local solutions are the local solutions of the (original) graph. 
	After the extraction of each cluster,  the constraint set is changed in such a way to make the extracted cluster unstable under the dynamics. The within-camera tracking starts with all nodes as constraint set. Let us say $\Gamma^i$ is the $i^{th}$ extracted cluster, $\Gamma^1$ is then the first extracted cluster which contains a subset of elements from the whole set.  After our first extraction, we change the constraint set to a set $V\backslash \Gamma^1$, hence rendering its associated nodes unstable (making the dynamics not able to select sets of nodes in the interior of associated nodes). The procedure iterates, updating the constraint set at the $i^{th}$ extraction as $V\backslash \bigcup\limits_{l=1}^i \Gamma^l$, until the constraint set becomes empty. Since we are not removing the nodes of the graph (after each extraction of a compact set), we may end up with a solution that assigns a node to more than one cluster.
	
	To find the final solution, we use the notion of centrality of constrained dominant sets. The true class of a node $j$, which is assigned to $\mat{K} > 1$ cluster, $\psi = \left\{\Gamma^1  \dots \Gamma^\mat{K} \right\}$, is  computed as:
	
	\[\argmax_{\Gamma^i \in \psi} ~ \left(|\Gamma^i|*\delta^i_j\right),\]
	where the cardinality $|\Gamma^i|$ is the number of nodes that forms the $i^{th}$ cluster and $\delta^i_j$ is the membership score of node $j$ obtained when assigned to cluster $\Gamma^i$. The normalization using the cardinality is important to avoid any unnatural bias to a smaller set.
	
	Algorithm (\ref{alg:Algorithm1}), putting the number of cameras under consideration ($\mathcal{I}$) to 1 and $\mathcal{Q}$  as short-tracklets(tracklets) in the first(second) layer, is used to determine constrained dominant sets which correspond to tracklet(track)  in the first (second) layer.

	\subsection{Across-Camera Tracking} \label{across-camera tracking}
	\subsubsection{Graph Representation of Tracks and the Payoff Function}
	Given tracks ($T_i^j $, that is, the $j^{th} $ track of camera $i$) of different cameras from previous step, we build graph $G'(V',E',w')$, where nodes represent tracks and their corresponding affinity matrix $\mat{A}$ depicts the similarity between tracks.

	Assuming we have $\mathcal{I}$ number of cameras and $ \mat{A}^{i\times j}$ represents the similarity among tracks of camera $i$ and $j$, the final track based affinity $\mat{A}$, is built as
	
	$$
	\mat{A} = 
	\begin{pmatrix} 
	\mat{A}^{1 \times 1} & .  .  & \mat{A}^{1 \times j} & . . & \mat{A}^{1 \times \mathcal{I}} \\ 
	. & ~~ . ~~ &  .  & . & .\\
	~~ ~~ & ~~ ~~ & ~~ ~~ &   \\
	\mat{A}^{i \times 1} & .  .  & \mat{A}^{i \times j} & . & \mat{A}^{i \times \mathcal{I}}\\ 
	. & ~~ ~~ &  . & . & .\\
	~~ ~~ & ~~ ~~ & ~~ ~~\\
	\mat{A}^{\mathcal{I} \times 1} & .  .  & \mat{A}^{\mathcal{I} \times j} & . . & \mat{A}^{\mathcal{I} \times \mathcal{I}}\\ 
	\end{pmatrix}.
	$$ 
	
\begin{figure}[!h]
		\centering
		\includegraphics[width=1\linewidth]{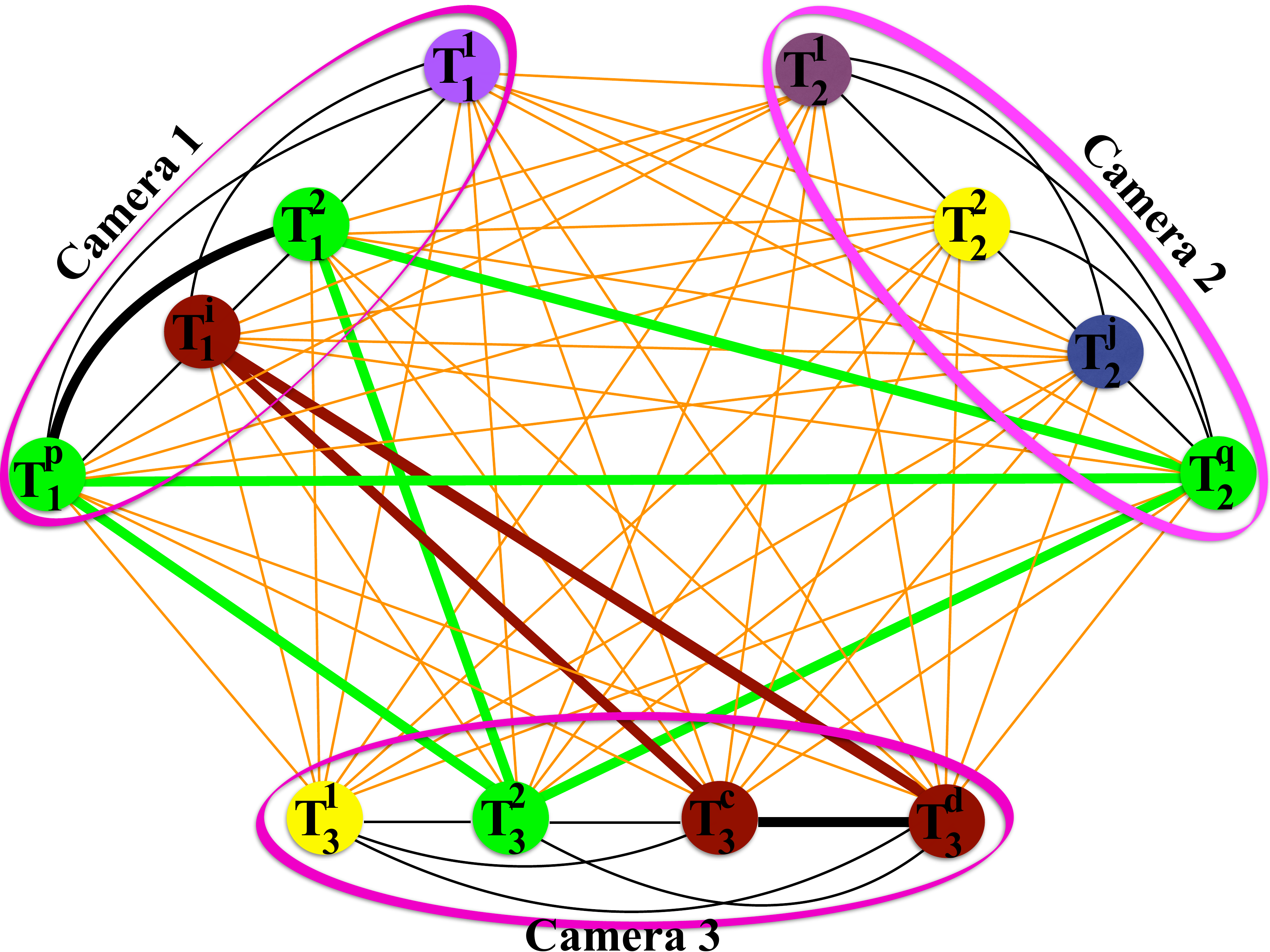}
		\caption{\small Exemplar graph of tracks from three cameras. $T^i_j$ represents the $i^{th}$ track of camera $j$.  Black and colored edges, respectively, represent within- and across-camera relations of tracks. Colours of the nodes depict track IDs, nodes with similar colour represent tracks of the same person, and the thick lines show both within- and across-camera association. 
		}
		\label{fig:ExamplarGraphWithinAndAcross}
	\end{figure}

	Figure \ref{fig:ExamplarGraphWithinAndAcross} shows exemplar graph for  across-camera tracking among three cameras.  $T^i_j$ represents the $i^{th}$ track of camera $j$. Black and orange edges, respectively, represent within- and across-camera relations of the tracks. From the affinity $\mat{A}$, $\mat{A}^{i\times j}$ represents the black edges of camera $i$ if $i = j$, which otherwise represents the across-camera relations using the orange edges. 
	
	The colors of the nodes depict the track ID; nodes with similar color represent tracks of the same person. Due to several reasons such as long occlusions, severe pose change of a person, reappearance and others, a person may have more than one track (a \textit{broken track}) within a camera. The green nodes of camera 1 (the second and the $p^{th}$ tracks) typify two \textit{broken tracks} of the same person, due to reappearance as shown in Figure \ref{fig:within_camera_2}. The proposed unified approach, as discussed in the next section, is able to deal with such cases. 
	
	\subsubsection{Across-camera Track Association}

	In this section, we discuss how we simultaneously solve within- and across-camera tracking.
	Our framework is  naturally able to deal with the  errors listed above. 
	A person, represented by the green node from our exemplar graph (Figure \ref{fig:ExamplarGraphWithinAndAcross}), has two tracks which are difficult to merge during within-camera tracking; however, they belong to clique (or cluster) with tracks in camera 2 and camera 3, since they  are highly similar. The algorithm applied to a such across-camera graph is able to cluster all the correct tracks. 
	This helps us linking \textit{broken tracks} of the same person occurring during within-camera track generation stage. 
	
	Using the graph with nodes of tracks from a camera as a constraint set, data association for both within- and across-camera are performed simultaneously. Let us assume, in our exemplar graph (Figure \ref{fig:ExamplarGraphWithinAndAcross}), our constraint set $\mathcal{Q}$ contains nodes of tracks of camera 1, $\mathcal{Q}$ = \{ $T^1_1, T^2_1, T^i_1, T^p_1$ \}.  $I_\mathcal{Q}$ is then $n \times n$ diagonal matrix, whose diagonal elements are  set to 1 in correspondence to the vertices contained in all cameras, except camera 1 which takes the value zero. That is,  the sub-matrix $I_\mathcal{Q}$, that corresponds to $A^{1\times 1}$, will be a zero matrix of size equal to number of tracks of the corresponding camera. Setting $\mathcal{Q}$ as above, we have guarantee that the maximizer of program in eq. (\ref{eqn:parQP}) contains some elements from set $\mathcal{Q}$: i.e., $\mathcal{C}_1^1$=$\left\{T^2_1, T^p_1, T^q_2, T^2_3\right\}$ forms a clique which contains set $\left\{T^2_1, T^p_1\right\} \in \mathcal {Q}$. 
	This is  shown in Figure \ref{fig:ExamplarGraphWithinAndAcross}, using the thick green edges (which illustrate across-camera track association) and the thick black edge (which typifies the within camera track association). The second set, $\mathcal{C}_1^2$, contains tracks shown with the dark red color, which illustrates the case where within- and across-camera tracks are in one clique. Lastly, the $\mathcal{C}_1^3$ = $T^1_1$ represents a track of a person that appears only in camera 1. 
	As a general case, $C^i_j$, represents the $i^{th}$ track set using tracks in camera $j$ as a constraint set and $C_j$ is the set that contains track sets generated using camera $j$ as a constraint set, e.g. $C_1 = \left\{ C^1_1, C^2_1, C^3_1\right\}$. We iteratively  process all the cameras and then apply track refinement step.
	
	Though Algorithm (\ref{alg:Algorithm1}) is applicable to within-camera tracking also, here we show the specific case for across-camera track association.
	Let $\mathcal{T}$ represents the set of tracks from all the cameras we have and $C$ is the set which contains sets of tracks, as $C^i_p$, generated using our  algorithm. $T^{\vartheta}_p$ typifies the $\vartheta^{th}$ track from camera $p$ and $T_p$ contains all the tracks in camera $p$. The function $\mathcal{F}(\mathcal{Q},\mat{A} $) takes as an input a constraint set $\mathcal{Q}$ and the affinity $\mat{A}$, and provides as output all the $m$ local solutions $\mathcal{X}^{n\times m}$ of program \eqref{eqn:parQP} that contain element(s) from the constraint set. This can be accomplished by iteratively finding a local maximizer of equation (program) (\ref{eqn:parQP}) in $\Delta$, e.g. using game dynamics, and then changing the constraint set $\mathcal{Q}$, until all members of the constraint set have been clustered.
	
	\begin{figure}[!h]
		\removelatexerror
		\begin{algorithm}[H]
			\caption{Track Association}
			{\bf INPUT:} Affinity $\mat{A}$, Sets of tracks $\mathcal{T}$ from $\mathcal{I}$ cameras\;
			$C$ $\leftarrow \emptyset$ Initialize the set with empty-set \;
			Initialize $\x$ to the barycenter and $i$ and $p$ to 1\;
			
			\While(){$p\le \mathcal{I}$}
			{
				$\mathcal{Q}$ $\leftarrow$ $T_p$, define constraint set\;
				$\mathcal{X}$ $\leftarrow$ $\mathcal{F}(\mathcal{Q},\mat{A} $)\;
				$C^i_p$ = $\leftarrow \sigma(\mathcal{X}^i)$, compute for all $i = 1 \dots m$\;
				$p$ $\leftarrow$ $p+1$\;			
			}
			$C$ = $\bigcup\limits_{p=1}^\mathcal{I} C_p$\;
			{\bf OUTPUT:} \{$C$\}
			\label{alg:Algorithm1}	
		\end{algorithm}
	\end{figure}
	
	\subsection{Track Refinement}
	
	The proposed framework, together with the notion of centrality of constrained dominant sets and the notion of reciprocal neighbours, helps us in refining tracking results  using tracks from different cameras as different constraint sets. Let us assume we have $\mathcal{I}$ cameras and $\mathcal{K}^i$ represents the set corresponding to track $i$,  while $\mathcal{K}^i_p$ is the subset of $\mathcal{K}^i$ that corresponds to the $p^{th}$ camera.  $\mathcal{M}^{l^i}_p$ is  the membership score assigned to the $l^{th}$ track in the set $\mathcal{C}^i_p$. 
	
	We use two constraints during 	track refinement stage, which  helps us refining false positive association.
	
	\noindent {\bf Constraint-1:}  {\em A track can not be found in two different sets generated using same constraint set}, i.e. it must hold that:
	
	\[|\mathcal{K}_p^i| \le 1 \] 
	
	Sets that do not satisfy the above inequality should be refined as there is one or more tracks that exist in different sets of tracks collected using the same constraint, i.e. $T_p$. The corresponding track is removed  from all the sets which contain it and is assigned to the right set based on its membership score in each of the sets. Let us say the $l^{th}$ track exists in $q$ different sets, when tracks from camera $p$ are taken as a constraint set, $|\mathcal{K}^l_p|=q$. The right set which contains the track, $C^r_p$, is chosen as:  
	
	\[C^r_p = \argmax_{C^i_p \in \mathcal{K}^l_p} ~ \left(|C^i_p|*\mathcal{M}^{l^i}_p \right). \]
	where $i = 1, \dots, |\mathcal{K}^l_p| $.
	This must be normalized with the cardinality of the set to avoid a bias towards smaller sets.
	
	\noindent {\bf Constraint-2:} {\em The maximum number of sets that contain track $i$ should be the number of cameras under consideration}. If we consider $\mathcal{I}$ cameras, the cardinality of the set which contains sets with track $i$, is not larger than $\mathcal{I}$, i.e.: 
	
	\[|\mathcal{K}^i| \le \mathcal{I}. \] 
	
	If there are sets that do not satisfy the above condition, the tracks are refined based on the cardinality of the intersection of sets that contain the track, i.e. by enforcing the reciprocal properties of the sets.
	
	If there are sets that do not satisfy the above condition, the tracks are refined based on the cardinality of the intersection of sets that contain the track by enforcing the reciprocal properties of the sets which contain a track. 
	Assume we collect sets of tracks considering tracks from camera $q$ as constraint set and assume a track $\vartheta$ in the set $C^j_p$, $p\ne q$, exists in more than one sets of $C_q$.  The right set, $C^r_q$, for $\vartheta$ considering tracks from camera $q$ as constraint set is chosen as:
	
	\[C^r_q = \argmax_{C^i_q \in \mathcal{K}^\vartheta_q} ~ \left( C^i_q \cap C^j_p \right).\] where $i = 1, \dots, |\mathcal{K}^\vartheta_q| $.

	\section{Fast Approach for Solving Constrained Dominant Set Clustering } \label{speedin up data association }

	Our constrained quadratic optimization program can be solved using dynamics from evolutionary game theory. The well-known standard game dynamics to equilibrium selection, replicator dynamics, though efficient, poses serious efficiency problems, since the time complexity for each iteration of the replicator dynamics is $\mathcal{O}(n^2)$, which makes it not efficient for large scale data sets \cite{ZemPelECCV16}. Rota Bul{\`{o}} \etal \cite{BulPelBomCVIU11} proposed a new class of evolutionary game dynamics, called Infection and Immunization Dynamics (InfImDyn).  InfImDyn solves the problem in linear time. However, it needs the whole affinity matrix  to extract a dominant set which, more often than not, exists in local range of the whole graph.
	Dominant Set Clustering (DSC) \cite{PavPelPAMI07} is an iterative method which, at each iteration, peels off a cluster by performing a replicator dynamics until its convergence. Efficient out-of-sample \cite{PavPelNIPS04}, extension of dominant sets, is the other approach which is used to reduce the computational cost by sampling the nodes of the graph using some given sampling rate that affects the framework efficacy. Liu \etal \cite{LiuLatYanPAMI13} proposed an iterative clustering algorithm, which operates in two steps: Shrink and Expansion. These steps help reduce the runtime of replicator dynamics on the whole data, which might be slow. The approach has many limitations such as its preference of  sparse graph with many small clusters and the results are sensitive to some additional parameters. 
	Another approach which tries to reduce the computational complexity of the standard quadratic program (StQP \cite{Bom02}) is proposed by \cite{ChuWanLiuHuaPeiPVLDB15}. 
	
	All the above formulations, with their limitations, try to minimize the computational complexity of StQP using the standard game dynamics, whose complexity is $\mathcal{O}(n^2)$ for each  iteration. 
	
	In this work we propose a fast approach (listed in Algorithm \ref{alg:Algorithm2}), based on InfImDyn approach which solves StQP in $\mathcal{O}(n)$, for the recently proposed formulation, $\x\T ( {\mat{A}} - \alpha  I_{\mathcal{Q}}) \x, $ which of-course generalizes the StQP.
	
	InfImDyn is a game dynamics inspired by Evolutionary game theory. The dynamics extracts a dominant set using a two-steps approach (infection and immunization), that iteratively increases the compactness measure of the objective function by driving the (probability) distribution with lower payoff to extinction, by determining an ineffective distribution $\vct{y}\in \Delta$, that satisfies the inequality $ (\vct y - \vct x)\T\mat{A}\vct x > 0$, the dynamics combines linearly
	the two distributions ($\vct{x}$ and $\vct{y}$), thereby engendering a new population $\vct{z}$ which is immune to $\vct{y}$ and guarantees a maximum increase in the expected payoff. In our setting, given a set of instances (tracks, tracklets) and their affinity, we first assign all of them an equal probability (a distribution at the centre of the simplex, a.k.a. barycenter). The dynamics then drives the initial distribution with lower affinity to extinction; those which have higher affinity start getting higher, while the other get lower values. A selective function, $\mathcal{S}(\vct{x})$, is then run to check if there is any infective distribution; a distribution which contains instances with a better association  score. By iterating this process of infection and immunization the dynamics is said to  reach the equilibrium, when the population is driven to a state that cannot be infected by any other distribution, that is there is no distribution, whose support contains a set of instances with a better association score. The selective function, however, needs whole affinity matrix, which makes the InfImDyn inefficient for large graphs. We  propose an algorithm, that reduces the search space using the Karush-Kuhn-Tucker (KKT) condition of the constrained quadratic optimization, effectively enforcing the user constraints. In the constrained optimization framework \cite{ZemPelECCV16}, the algorithm computes the eigenvalue of the submatrix for every extraction of the compact sets, which contains the user constraint set. Computing eigenvalues for large graphs is computationally intensive, which makes the whole algorithm inefficient. 
	
	In our approach, instead of running the dynamics over the whole graph, we localize it on the sub-matrix, selected using the dominant distribution, that is much smaller than the original one. To alleviate the issue with the eigenvalues, we utilize the properties of eigenvalues; a good approximation for the parameter $\alpha$ is to use the maximum degree of the graph, which of-course is larger than the eigenvalue of corresponding matrix. The computational complexity, apart from eigenvalue computation, is reduced to $\mathcal{O}(r)$ where $r$, which is much smaller than the original affinity,  is the size of the sub-matrix where the dynamics is run.

	Let us summarize the KKT conditions for quadratic program reported in eq. (\ref{eqn:parQP}). By adding Lagrangian multipliers, $n$ non-negative constants $ \mu_1, ...., \mu_n$ and a real number $\lambda$, its Lagrangian function is defined as follows: 
	
	\begin{displaymath}
		\mathcal{L}(x,\mu,\lambda) = f_{\mathcal{Q}}^\alpha(\x) + 
		\lambda \left(1 - \sum\limits_{i+1}^{n}x_i \right) +
		\sum\limits_{i+1}^{n}\mu_ix_i.
	\end{displaymath}
	
	For a distribution $x \in \Delta$ to be a KKT-point, in order to satisfy the first-order necessary conditions for local optimality \cite{Lue84}, it should satisfy the following two conditions:
	
	\begin{displaymath}
		2*[(A - \alpha I_{\mathcal{Q}}) \x]_i - \lambda + \mu_i = 0,
	\end{displaymath}
	for all $i=1 \ldots n$, and
	\begin{displaymath}
		\sum_{i=1}^n x_i \mu_i = 0~.
	\end{displaymath}
	Since both the $x_i$ and the $\mu_i$ values are nonnegative, the
	latter condition is equivalent to saying that $i \in \sigma(\x)$ which implies that 
	$\mu_i= 0$, from which we obtain:
	\begin{equation}
		\label{KKT}
		[(A - \alpha I_{\mathcal{Q}}) \x]_i ~
		\begin{cases} 
			~ = ~   \lambda/2, ~ \mbox{ if } i \in \sigma(\x) \\ 
			~ \le ~ \lambda/2, ~ \mbox{ if } i \notin \sigma(\x)  
		\end{cases}
	\end{equation}

	We then need to define a \textit{Dominant distribution}
	\begin{Definition}
		\label{def:dominantDistribution}
		A distribution $\y \in \Delta$ is said to be a \textbf{dominant distribution} for $\x \in \Delta$ if 
		
		\begin{equation}
			\left\{\sum\limits_{i,j=1}^{n}   x_iy_ja_{ij} - \alpha x_iy_j \right\} > \left\{\sum\limits_{i,j=1}^{n}   x_ix_ja_{ij} - \alpha x_ix_j\right\}
		\end{equation}
	\end{Definition}
		
	Let the "support" be
	$
	\sigma(\x) = \{i \in V~:~x_i > 0\}
	$ 
	and  $e_i$ the $i^{th}$ unit vector (a zero vector whose $i^{th}$ element is one).	
	
	\begin{proposition}
		Given an affinity A and a distribution $\x \in \Delta$, if $(A\x)_i             > \x'A\x - \alpha \x'_{\mathcal{Q}} \x_{\mathcal{Q}}, \mbox{for} \;i \notin \sigma(\x)$,	
		\begin{enumerate}
			\item $\x$ is not the maximizer of the parametrized quadratic program of (\ref{eqn:parQP})
			\item $e_i$ is a \textbf{dominant distribution} for $\x$
		\end{enumerate}	
	\end{proposition}
	(We refer the reader to appendix for the proof)
	%
	%
	%
	%
	%
	%

	The proposition provides us with an easy-to-compute dominant distribution.

	Let a function, $\mathcal{S}(\mat{A},x)$, returns a dominant distribution for distribution, $x$, $\emptyset$ otherwise and $\mathcal{G}(\mat{A},\mathcal{Q},x)$ returns the local maximizer of program \eqref{eqn:parQP}. We summarize the details of our proposed algorithm in Algorithm (\ref{alg:Algorithm2})
	\begin{figure}[!h]
		\removelatexerror
		\begin{algorithm}[H]
			\caption{Fast CDSC}
			{\bf INPUT:} Affinity $\mat{B}$, Constraint set $\mathcal{Q}$\;
			Initialize $\x$ to the barycenter of $\Delta_{\mathcal{Q}}$\;
			$\x_d$ $\leftarrow$ $\x$, initialize \textit{dominant distribution} \;
			\While(){true}
			{
				
				$\x_d$ $\leftarrow$ $\mathcal{S}(\mat{B},\x)$, Find dominant distribution for $x$ \;
				if $\x_d = \emptyset$ break \; 
				$\mathcal{H}$ $\leftarrow$ $\sigma(\x_d) \cup \mathcal{Q}$, subgraph nodes\;
				$\mat{A}\leftarrow \mat{B}_{\mathcal{H}}$\;
				$\x_l$ $\leftarrow$ $\mathcal{G}(\mat{A},\mathcal{Q},x)$\;
				$\x$ $\leftarrow$ $\x$*0\;
				$\x (\mathcal{H})$ $\leftarrow$ $\x_l$\;
			}
			
			{\bf OUTPUT:} \{\x\}
			\label{alg:Algorithm2}	
		\end{algorithm}
	\end{figure}
	
	The selected dominant distribution always increases the value of the objective function. Moreover, the objective function is bounded  which guaranties the convergence of the algorithm.
	\section{Experimental Results} \label{experiments}
	The proposed framework has been evaluated on recently-released large dataset, MOTchallenge DukeMTMC \cite{RisSolZouCucTomECCV16,solera2016groups,ErgCarACCV2014}. Even though the main focus of this paper is on multi-target tracking in multiple non-overlapping cameras, we also perform additional experiments on MARS  \cite{ZheBieSunWanSuWanTiaECCV16}, one of the largest and challenging video-based person re-identification dataset, to show that the proposed across-camera tracking approach can efficiently solve this task also.  
	
	\textbf{DukeMTMC}  is recently-released dataset to evaluate the performance of multi-target multi-camera tracking systems. It is the largest (to date),  fully-annotated and calibrated high resolution 1080p, 60fps dataset, that covers a single outdoor scene from 8 fixed synchronized cameras, the topology of cameras is shown in Fig. 4. The dataset consists of 8 videos of 85 minutes each from the 8 cameras, with 2,700 unique identities (IDs) in more than 2 millions frames in each video  containing 0 to 54 people. The video is split in three parts: (1) Trainval (first 50 minutes of the video), which is for training and validation; (2) Test-Hard (next 10 minutes after Trainval sequence); and (3) Test-Easy, which covers the last 25 minutes of the video. Some of the properties which make the dataset more challenging include: huge amount of data to process, it contains 4,159 hand-overs, there are more than 1,800 self-occlusions (with 50\% or more overlap),  891 people walking in front of only one camera.
	\begin{figure}[h]
		\centering 
		\includegraphics[width=1\linewidth, scale = 0.4 ,trim=0cm 0cm 0cm 0cm,clip]{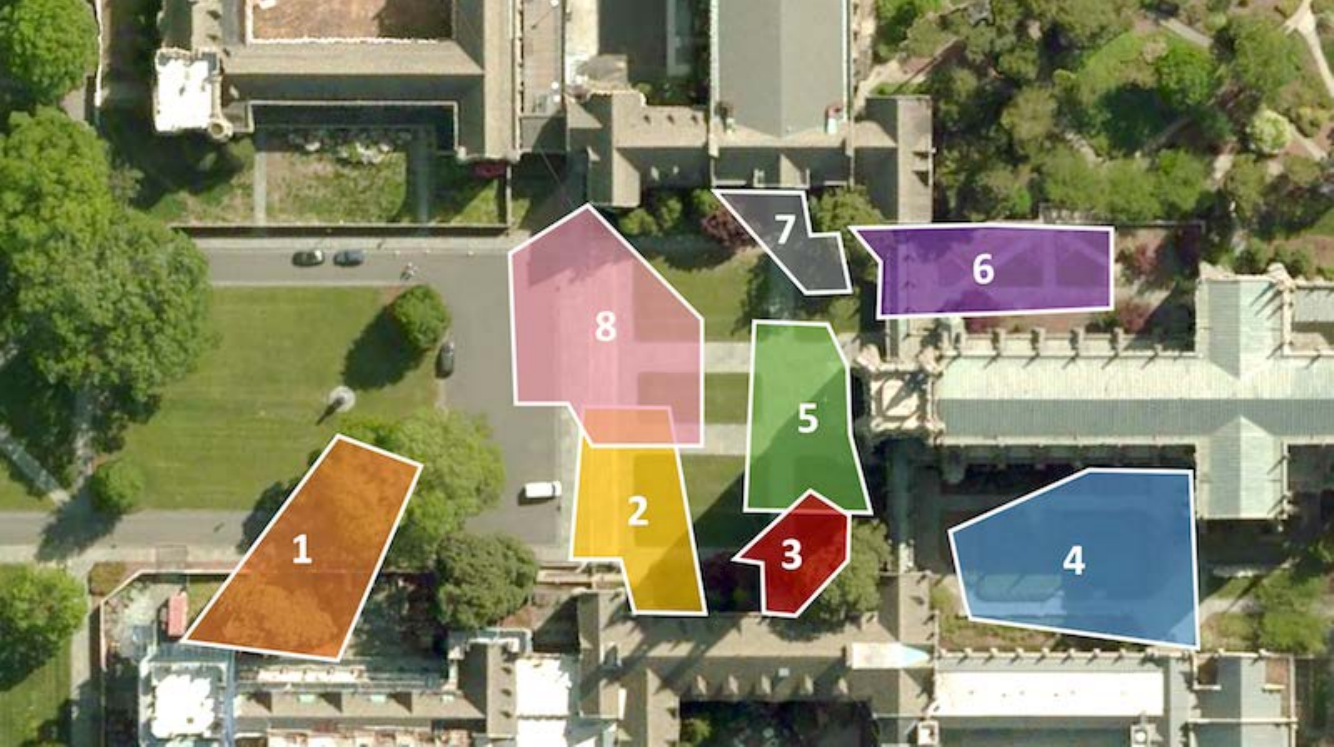}
		\caption{ Camera topology for DukeMTMC dataset. Detections from the overlapping fields of view are not considered. More specifically, intersection occurred between camera (8 \& 2) and camera (5 \& 3). }
		\label{fig:Camera_topology}
	\end{figure}
	
	\textbf{MARS (Motion Analysis and Re-identification Set)}  is an extension of the Market-1501 dataset \cite{ZheBieSunWanSuWanTiaECCV16}. It has been collected from six near-synchronized cameras. It consists of 1,261 different pedestrians, who are captured by at least 2 cameras. The variations in poses, colors and illuminations of pedestrians, as well as the poor image quality, make it very  difficult to yield high matching accuracy. Moreover, the dataset contains 3,248 distractors in order to make it more realistic. Deformable Part Model (DPM) \cite{FelGirMcaRamPAMI10} and GMMCP tracker \cite{gmmcp} were used to automatically generate the tracklets (mostly 25-50 frames long). Since the video and the detections are not available we use the generated tracklets as an input to our framework.

	\textbf{Performance Measures:} In addition to the standard Multi-Target Multi-Camera tracking performance measures, we evaluate our framework using additional measures recently proposed in \cite{RisSolZouCucTomECCV16}: Identification F-measure (IDF1), Identification Precision (IDP) and Identification Recall (IDR) \cite{RisSolZouCucTomECCV16}. The standard performance measures such as CLEAR MOT report the amount of  incorrect decisions made by a tracker. Ristani \etal \cite{RisSolZouCucTomECCV16} argue and demonstrate that some system users may instead be more interested in how well they can determine who is where at all times. After pointing out that different measures serve different purposes, they proposed the three measures (IDF1, IDP and IDR) which can be applied both within- and across-cameras. These measure tracker's performance not by how often ID switches occur, but by how long the tracker correctly tracks targets.
	
	\textbf{\textit{Identification precision IDP (recall IDR)}:} is the fraction of computed (ground truth) detections that are correctly identified. 
	
	\textbf{\textit{Identification F-Score IDF1}:} is the ratio of correctly identified detections over the average number of ground-truth and computed detections. Since MOTA and its related performance measures under-report across-camera errors \cite{RisSolZouCucTomECCV16}, we use them for the evaluation of our single camera tracking results.
	
	The performance of the algorithm for re-identification is evaluated employing rank-1 based accuracy  and confusion matrix using average precision (AP).

	\textbf{Implementation:} In the implementation of our framework, we do not have parameters to tune. The affinity matrix $\mat{A}$ adapting kernel trick distance function from \cite{GroBayBenACM05}, is constructed as follows:
	
	\[ \mat{A}_{i,j} = 1 - \sqrt{\frac{\mat{K}(x_i,x_i)+\mat{K}(x_j,x_j) - 2*\mat{K}(x_i,x_j)} {2}},\]
	where $\mat{K}(x_i,x_j)$ is chosen as the Laplacian kernel 
	
	\[ exp(-\gamma\parallel x_i - x_j \parallel _1).\]
	The kernel parameter $\gamma$ is set as the inverse of the median of pairwise distances.
	
	In our similarity matrix for the final layer of the framework, which is sparse, we use spatio-temporal information based on the time duration and the zone of a person moving from one zone of a camera to other zone of another camera which is learned from the Trainval sequnece of DukeMTMC dataset. The affinity between track $i$ and track $j$ is different from zero , if and only if they have a possibility, based on the direction a person is moving and the spatio-temporal information, to be linked and form a trajectory (across camera tracks of a person). However, this may have a drawback due to \textit{broken tracks} or track of a person who is standing and talking or doing other things in one camera which results in a track that does not meet the spatio-temporal constraints. To deal with this problem, we add, for the across camera track's similarity, a path-based information as used in \cite{EyaMarICIAP2015}, i.e if a track in camera $i$ and a track in camera $j$ have a probability to form a trajectory,  and track $j$ in turn have linkage possibility with a track in camera $z$, the tracks in camera $i$ and camera $z$ are considered to have a possibility to be linked.
	
	The similarity between two tracks is computed using the Euclidean distance of the max-pooled features. The max-pooled features  are computed as the row maximum of the feature vector of individual patch, of the given track, extracted from the last fully-connected layer of Imagenet pre-trained 50-layers Residual Network (ResNet\_50) \cite{KaiXiaShaJiaCVPR2016}, fine-tuned using the Trainval sequence of DukeMTMC dataset. The network is fine-tuned  with classification loss on the Trainval sequence, and activations of its last fully-connected layer are extracted, L2-normalized and taken as visual features. Cross-view Quadratic Discriminant Analysis (XQDA) \cite{LiaHuZhuLiCVPR15} is then used for pairwise distance computation between instances. For the experiments on MARS, patch representation is obtained using CNN features used in \cite{ZheBieSunWanSuWanTiaECCV16}. The pairwise distances between instances are then computed in XQDA, KISSME \cite{KosHirWohRotBisCVPR12} and euclidean spaces.
	

	\begin{table*}[t]
		\centering	
		\begin{tabular}{cc|c|c|c|c|c|c|c|c|c|c|r}
			\centering
			& {Methods} &{MOTA$\uparrow$} & {MOTP$\uparrow$} &{FAF$\downarrow$}&{MT$\uparrow$} & {ML$\downarrow$}  &{FP$\downarrow$} & {FN$\downarrow$} &{IDS$\downarrow$}  &{IDF1}$\uparrow$ & IDP$\uparrow$ &  IDR$\uparrow$ \\ \cline{1-13}
			\cline{1-13}
			\multicolumn{1}{ c  }{\multirow{2}{*}{Camera1} } &
			\multicolumn{1}{ |c| }{\cite{RisSolZouCucTomECCV16}} & 43.0	& 79.0&	0.03& 24&46&2,713&107,178&39 &57.3&91.2&41.8   \\ \cline{2-13}
			\multicolumn{1}{ c  }{}                        &
			\multicolumn{1}{ |c| }{Ours} &  69.9  & 76.3 & 0.06 & 137& 22 & 5,809 & 52,152 & 156 & 76.9 & 89.1 & 67.7 \\ \cline{1-13}
			\multicolumn{1}{ c  }{\multirow{2}{*}{Camera2} } &
			\multicolumn{1}{ |c| }{\cite{RisSolZouCucTomECCV16}} & 44.8&	78.2&	0.51 & 133 & 8 & 47,919 & 53,74 & 60 & 68.2 &	69.3 &67.1   \\ \cline{2-13}
			\multicolumn{1}{ c  }{}                        &
			\multicolumn{1}{ |c| }{Ours} & 71.5 & 74.6 & 0.09 & 134 & 21 & 8,487 & 43,912 & 75 & 81.2 & 90.9 & 73.4  \\ \cline{1-13}
			\multicolumn{1}{ c  }{\multirow{2}{*}{Camera3} } &
			\multicolumn{1}{ |c| }{\cite{RisSolZouCucTomECCV16}} & 	57.8&	77.5&	0.02 &	52 &	22 &	1,438 & 28,692 &	16 & 60.3&	78.9&	48.8   \\ \cline{2-13}
			\multicolumn{1}{ c  }{}                        &
			\multicolumn{1}{ |c| }{Ours} & 67.4 & 75.6 & 0.02 & 44 & 9 & 2,148 & 21,125 & 38 & 64.6 & 76.3 & 56.0 \\ \cline{1-13}
			\multicolumn{1}{ c  }{\multirow{2}{*}{Camera4} } &
			\multicolumn{1}{ |c| }{\cite{RisSolZouCucTomECCV16}} & 63.2&	80.2&	0.02&	36&	18&	2,209&	19,323&	7 & 	73.5&	88.7&	62.8   \\ \cline{2-13}
			\multicolumn{1}{ c  }{}                        &
			\multicolumn{1}{ |c| }{Ours} & 76.8 & 76.6 & 0.03 & 45 & 4 & 2,860 & 10,689 & 18 & 84.7 & 91.2 & 79.0   \\ \cline{1-13}
			
			\multicolumn{1}{ c  }{\multirow{2}{*}{Camera5} } &
			\multicolumn{1}{ |c| }{\cite{RisSolZouCucTomECCV16}} & 72.8 &	80.4&	0.05&	107&17&	4,464&35,861&	54&73.2&	83.0&	65.4    \\ \cline{2-13}
			\multicolumn{1}{ c  }{}                        &
			\multicolumn{1}{ |c| }{Ours} & 68.9 & 77.4 & 0.10 & 88 & 11 & 9,117 & 36,933 & 139 & 68.3& 76.1 & 61.9   \\ \cline{1-13}
			\multicolumn{1}{ c  }{\multirow{2}{*}{Camera6} } &
			\multicolumn{1}{ |c| }{\cite{RisSolZouCucTomECCV16}} & 73.4&	80.2&	0.06&	142&27&	5,279&45,170& 55&	77.2&	87.5&	69.1   \\ \cline{2-13}
			\multicolumn{1}{ c  }{}                        &
			\multicolumn{1}{ |c| }{Ours} & 77.0 & 77.2 & 0.05 & 136 & 11 & 4,868 & 38,611 & 142 & 82.7 & 91.6 &75.3  \\ \cline{1-13}
			\multicolumn{1}{ c  }{\multirow{2}{*}{Camera7} } &
			\multicolumn{1}{ |c| }{\cite{RisSolZouCucTomECCV16}} &	71.4&	74.7&	0.02&	69& 13&1,395&18,904&	23 & 80.5&	93.6&	70.6    \\ \cline{2-13}
			\multicolumn{1}{ c  }{}                        &
			\multicolumn{1}{ |c| }{Ours} & 73.8 & 74.0 & 0.01 & 64 & 4 & 1,182 & 17,411 & 36 & 81.8 & 94.0& 72.5    \\ \cline{1-13}
			\multicolumn{1}{ c  }{\multirow{2}{*}{Camera8} } &
			\multicolumn{1}{ |c| }{\cite{RisSolZouCucTomECCV16}} & 60.7&	76.7&	0.03&	102&53&2,730&52,806&	46&	 72.4&	92.2&	59.6     \\ \cline{2-13}
			\multicolumn{1}{ c  }{}                        &
			\multicolumn{1}{ |c| }{Ours} & 63.4 & 73.6 & 0.04 & 92 & 28 & 4,184 & 47,565 & 91 & 73.0& 89.1 & 61.0  \\ \cline{1-13}
			\multicolumn{1}{ c  }{\multirow{2}{*}{Average} } &
			\multicolumn{1}{ |c| }{\cite{RisSolZouCucTomECCV16}} & 59.4&\textbf{78.7}&0.09& 665&234&68,147&361,672&\textbf{300}& 70.1 &83.6&	60.4    \\ \cline{2-13}
			\multicolumn{1}{ c  }{}                        &
			\multicolumn{1}{ |c| }{Ours} &  \textbf{70.9}  &75.8 &\textbf{0.05} & \textbf{740} &\textbf{110} & \textbf{38,655} &  \textbf{268,398} & 693 & \textbf{77.0} & \textbf{87.6} &   \textbf{68.6}\\ \cline{1-13}
			
		\end{tabular}
		
		\caption{The results show detailed (for each camera) and average performance of our and state-of-the-art approach \cite{RisSolZouCucTomECCV16} on the Test-Easy sequence of DukeMTMC dataset.}
		\label{table:TestEasyDuke}
	\end{table*}
	
	\begin{table*}[h]
		\centering

		\begin{tabular}{cc|c|c|c|c|c|c|c|c|c|c|r}
			\centering
			& {Methods} &{MOTA$\uparrow$} & {MOTP$\uparrow$} &{FAF$\downarrow$}&{MT$\uparrow$} & {ML$\downarrow$}  &{FP$\downarrow$} & {FN$\downarrow$} &{IDS$\downarrow$}  &{IDF1}$\uparrow$ & IDP$\uparrow$ &  IDR$\uparrow$ \\ \cline{3-13}
			\cline{1-13}
			\multicolumn{1}{ c  }{\multirow{2}{*}{Camera1} } &
			\multicolumn{1}{ |c| }{\cite{RisSolZouCucTomECCV16}} &  37.8	& 78.1	& 0.03	& 6	  & 34  & 1,257	      & 78,977	& 55	           & 52.7     & 92.5	  & 36.8     \\ \cline{2-13}
			\multicolumn{1}{ c  }{}                        &
			\multicolumn{1}{ |c| }{Ours} &   63.2 & 75.7 & 0.08 & 65 & 17 & 2,886 & 44,253 & 408 & 67.1 & 83.0 & 56.4 \\ \cline{1-13}
			\multicolumn{1}{ c  }{\multirow{2}{*}{Camera2} } &
			\multicolumn{1}{ |c| }{\cite{RisSolZouCucTomECCV16}} &  47.3	& 76.5	& 0.74	& 68  & 12   & 26526    & 46898	& 194        & 60.6	    & 65.7	  & 56.1      \\ \cline{2-13}
			\multicolumn{1}{ c  }{}                        &
			\multicolumn{1}{ |c| }{Ours} & 54.8 & 73.9 & 0.24 & 62 & 16 & 8,653 & 54,252 & 323 & 63.4 & 78.8 & 53.1 \\ \cline{1-13}
			\multicolumn{1}{ c  }{\multirow{2}{*}{Camera3} } &
			\multicolumn{1}{ |c| }{\cite{RisSolZouCucTomECCV16}} &  46.7	& 77.9	& 0.01	& 24	  & 4	    & 288	      & 18182	& 6	             & 62.7	    & 96.1	  & 46.5	  \\ \cline{2-13}
			\multicolumn{1}{ c  }{}                        &
			\multicolumn{1}{ |c| }{Ours} & 68.8 & 75.1 & 0.06 & 18 & 2 & 2,093 & 8,701 & 11 & 81.5 & 91.1 & 73.7 \\ \cline{1-13}
			\multicolumn{1}{ c  }{\multirow{2}{*}{Camera4} } &
			\multicolumn{1}{ |c| }{\cite{RisSolZouCucTomECCV16}} &  85.3	& 81.5	& 0.04	& 21	  & 0	    & 1,215	      & 2,073	& 1	              & 84.3   & 86.0	  & 82.7    \\ \cline{2-13}
			\multicolumn{1}{ c  }{}                        &
			\multicolumn{1}{ |c| }{Ours} & 75.6 & 77.7 & 0.05 & 17 & 0 & 1,571 & 3,888 & 61 & 82.3 & 87.1 & 78.1 \\ \cline{1-13}
			
			\multicolumn{1}{ c  }{\multirow{2}{*}{Camera5} } &
			\multicolumn{1}{ |c| }{\cite{RisSolZouCucTomECCV16}} &  78.3	& 80.7	& 0.04	& 57	  & 2	    & 1,480	      & 11,568	& 13	             & 81.9	    & 90.1	  & 75.1    \\ \cline{2-13}
			\multicolumn{1}{ c  }{}                        &
			\multicolumn{1}{ |c| }{Ours} & 78.6 & 76.7 & 0.03 & 47 & 2 & 1,219 & 11,644 & 50& 82.8 & 91.5 & 75.7   \\ \cline{1-13}
			\multicolumn{1}{ c  }{\multirow{2}{*}{Camera6} } &
			\multicolumn{1}{ |c| }{\cite{RisSolZouCucTomECCV16}} &  59.4	& 76.7	& 0.14	& 85	  & 23 & 5,156	      & 77,031	& 225      & 64.1	    & 81.7	  & 52.7      \\ \cline{2-13}
			\multicolumn{1}{ c  }{}                        &
			\multicolumn{1}{ |c| }{Ours} & 53.3& 76.5 & 0.17 & 68 & 36 & 5,989 & 88,164 & 547 & 53.1 & 71.2 & 42.3  \\ \cline{1-13}
			\multicolumn{1}{ c  }{\multirow{2}{*}{Camera7} } &
			\multicolumn{1}{ |c| }{\cite{RisSolZouCucTomECCV16}} &  50.8	& 73.3	& 0.08	& 43	  & 23  & 2,971	      & 38,912	& 148       & 59.6	    & 81.2	  & 47.1     \\ \cline{2-13}
			\multicolumn{1}{ c  }{}                        &
			\multicolumn{1}{ |c| }{Ours} & 50.8 & 74.0 & 0.05 & 34 & 20 & 1,935 & 39,865 & 266 & 60.6 & 84.7 & 47.1   \\ \cline{1-13}
			\multicolumn{1}{ c  }{\multirow{2}{*}{Camera8} } &
			\multicolumn{1}{ |c| }{\cite{RisSolZouCucTomECCV16}} &  73.0	& 75.9	& 0.02	& 34	  & 5	     & 706	      & 9735	& 10	            & 82.4   &94.9	  & 72.8     \\ \cline{2-13}
			\multicolumn{1}{ c  }{}                        &
			\multicolumn{1}{ |c| }{Ours} & 70.0 & 72.6 & 0.06 & 37 & 6 & 2,297 & 9,306 & 26 & 81.3 & 90.3 & 73.9   \\ \cline{1-13}
			\multicolumn{1}{ c  }{\multirow{2}{*}{Average} } &
			\multicolumn{1}{ |c| }{\cite{RisSolZouCucTomECCV16}} & 54.6	&\textbf{ 77.1	}& 0.14	& 338	&103	& 39,599	& 283,376	& \textbf{652}	   &64.5	& 81.2	& 53.5     \\ \cline{2-13}
			\multicolumn{1}{ c  }{}                        &
			\multicolumn{1}{ |c| }{Ours} & \textbf{59.6}& 75.4 & \textbf{0.09} & \textbf{348} & \textbf{99}& \textbf{26,643} & \textbf{260,073} & 1637 & \textbf{65.4} & \textbf{81.4}& \textbf{54.7}  \\ \cline{1-13}
			
		\end{tabular}
		
		\caption{The results show detailed (for each camera) and average performance of our and state-of-the-art approach \cite{RisSolZouCucTomECCV16} on the Test-Hard sequence of DukeMTMC dataset.}
		\label{table:TestHardDuke}
	\end{table*}
	
	\begin{table}[h]
		\centering
		\begin{tabular}{cc|c|c|r}
			
			& Methods  &{IDF1}$\uparrow$ & IDP$\uparrow$ &  IDR$\uparrow$ \\ \cline{3-4}
			\cline{1-5}
			\multicolumn{1}{ c  }{\multirow{2}{*}{Multi-Camera} } &
			\multicolumn{1}{ |c| }{\cite{RisSolZouCucTomECCV16}} & 
			56.2 &	67.0 &	48.4 \\ \cline{2-5}
			\multicolumn{1}{ c  }{}                        &
			\multicolumn{1}{ |c| }{Ours} & \textbf{60.0} & \textbf{68.3} &  \textbf{53.5}   \\ \cline{1-5}
			
		\end{tabular}
		
		\caption{Multi-camera performance of our and state-of-the-art approach \cite{RisSolZouCucTomECCV16} on the Test-Easy sequence of DukeMTMC dataset.}
		\label{table:MCTestEasyDuke}
	\end{table}

	\begin{table}[h]	
		\centering
		\begin{tabular}{cc|c|c|r}
			
			& Methods  &{IDF1}$\uparrow$ & IDP$\uparrow$ &  IDR$\uparrow$ \\ \cline{3-4}
			\cline{1-5}
			\multicolumn{1}{ c  }{\multirow{2}{*}{Multi-Camera} } &
			\multicolumn{1}{ |c| }{\cite{RisSolZouCucTomECCV16}} & 47.3	& 59.6 &39.2
			\\ \cline{2-5}
			\multicolumn{1}{ c  }{}                        &
			\multicolumn{1}{ |c| }{Ours} & \textbf{50.9} & \textbf{63.2} &   \textbf{42.6}  \\ \cline{1-5}
			
		\end{tabular}
		
		\caption{Multi-Camera performance of our and state-of-the-art approach \cite{RisSolZouCucTomECCV16} on the Test-Hard sequence of DukeMTMC dataset.}
		\label{table:MCTestHardDuke}
	\end{table}
	\subsection{Evaluation on DukeMTMC dataset:} In Table \ref{table:TestEasyDuke} and Table \ref{table:TestHardDuke}, we compare quantitative performance of our method with state-of-the-art multi-camera multi-target tracking method on the DukeMTMC dataset. The symbol $\uparrow$ means higher scores indicate better performance, while $\downarrow$ means lower scores indicate better performance. The quantitative results of the trackers shown in table \ref{table:TestEasyDuke} represent the performance on the  Test-Easy sequence, while those in table \ref{table:TestHardDuke} show the performance on the  Test-Hard sequence. For a fair comparison, we use the same detection responses obtained from MOTchallenge DukeMTMC as the input to our method. In both cases, the reported results of row 'Camera 1' to 'Camera 8' represent the within-camera tracking performances. The last row of the tables represent the average performance over 8 cameras. Both tabular results demonstrate that the proposed approach improves tracking performance for both sequences. In the Test-Easy sequence, the performance is improved by 11.5\% in MOTA and 7\% in IDF1 metrics, while in that of the Test-Hard sequence, our method produces 5\% larger average MOTA score than \cite{RisSolZouCucTomECCV16}, and 1\% improvement is achieved in IDF1. Table \ref{table:MCTestEasyDuke} and Table \ref{table:MCTestHardDuke} respectively present Multi-Camera performance of our and state-of-the-art approach \cite{RisSolZouCucTomECCV16} on the Test-Easy and  Test-Hard sequence (respectively) of DukeMTMC dataset. We have improved IDF1 for both Test-Easy and Test-Hard sequences by 4\% and 3\%, respectively.
	
	Figure \ref{fig:QualitativeResults} depicts sample qualitative results. Each person is represented by (similar color of) two bounding boxes, which represent the person's position at some specific time, and a track which shows the path s(he) follows. In the first row, all the four targets, even under significant illumination and pose changes, are successfully tracked in four cameras, where they appear. In the second row, target 714 is successfully tracked through three cameras. Observe its significant illumination and pose changes from camera 5 to camera 7. In the third row, targets that move through camera 1, target six, seven and eight are tracked. The last row shows tracks of targets that appear in cameras 1 to 4.

	\begin{figure}[h!]
		\centering 
		\includegraphics[width=1\linewidth ,trim=0cm 0cm 0cm 0cm,clip]{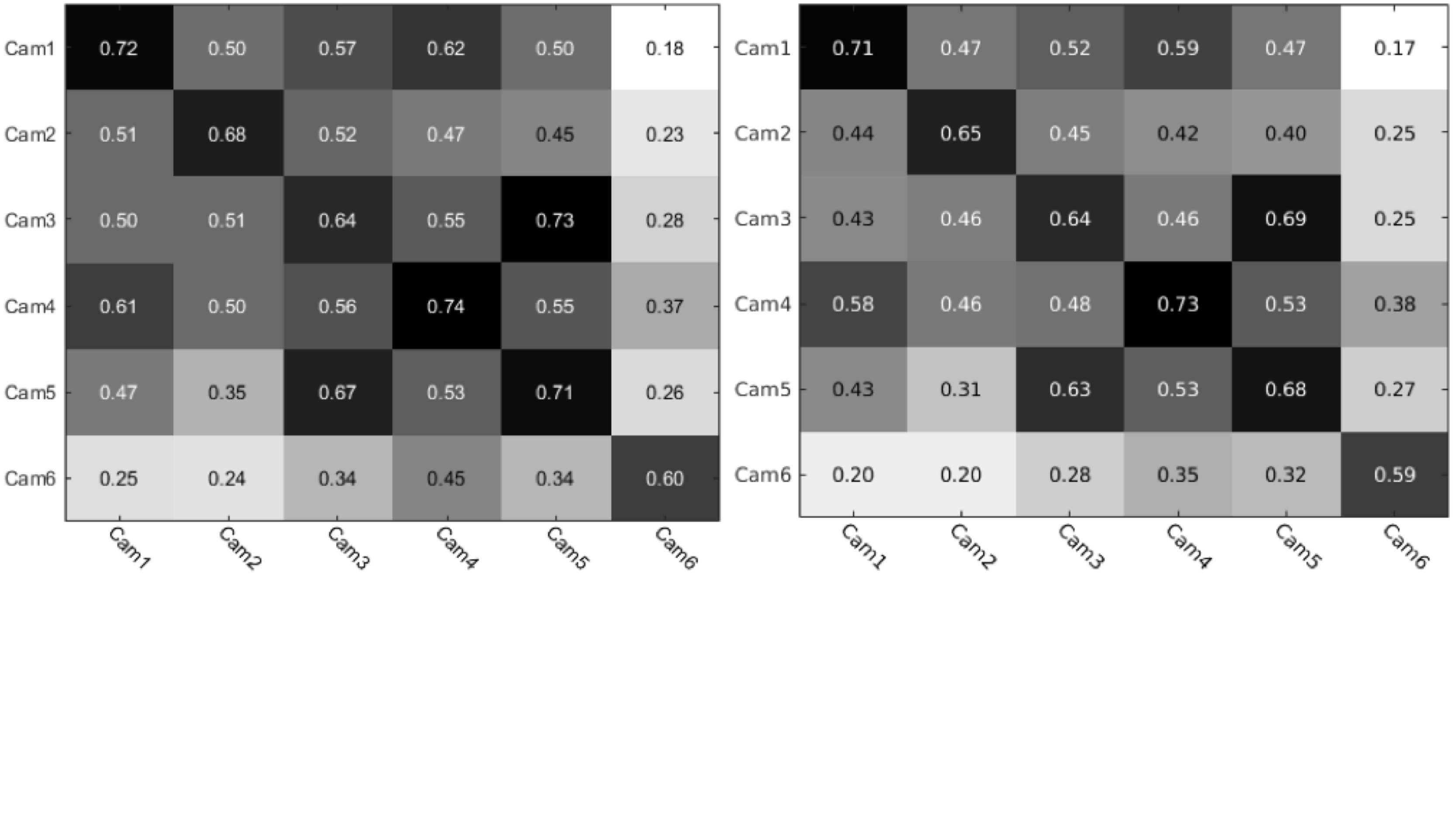}
		\vspace*{-2cm}
		\caption{The results show the performance of our algorithm on MARS (both using CNN + XQDA) when the final ranking is done using membership score (\textbf{left}) and using pairwise euclidean distance (\textbf{right}). 
		}
		\label{fig:NotionOfMembershipScore1}
	\end{figure}
	
	
	\subsection{Evaluation on MARS dataset:} In Table \ref{table:rank1} we compare our results (using the same settings as in \cite{ZheBieSunWanSuWanTiaECCV16}) on MARS dataset with the state-of-the-art methods.  The proposed approach achieves  3\% improvement. In table \ref{table:NotionOfMembershipScore} the results show performance of our  and state-of-the-art approach \cite{ZheBieSunWanSuWanTiaECCV16} in solving the within- (average of the diagonal of the confusion matrix, Fig. \ref{fig:NotionOfMembershipScore1}) and across-camera (off-diagonal average) ReID using average precision. Our approach shows up to 10\% improvement in the across-camera ReID and up to 6\% improvement in the within camera ReID.  
	
	\begin{table}[h!]
		{
			\centering
			\begin{tabular}{l|r}
				Methods                          & rank 1 \\ \hline
				HLBP + XQDA             & 18.60 \\ \hline
				BCov + XQDA              & 9.20 \\ \hline
				LOMO + XQDA                   & 30.70 \\ \hline
				BoW + KISSME            & 30.60 \\ \hline
				SDALF + DVR     & 4.10 \\ \hline
				HOG3D + KISSME    & 2.60 \\ \hline
				CNN + XQDA \cite{ZheBieSunWanSuWanTiaECCV16}   & 65.30 \\ \hline
				CNN + KISSME \cite{ZheBieSunWanSuWanTiaECCV16}   & 65.00 \\ \hline
				Ours   & \textbf{68.22} \\ \hline
			\end{tabular}
			\caption{The table shows the comparison (based on rank-1 accuracy) of our approach  with the state-of-the-art approaches: SDALF \cite{FarBazPerMurCriCVPR10}, HLBP \cite{XioGouCamSznECCV14}, BoW \cite{ZheSheTiaWanWanTiaICCV15}, BCov \cite{MaSuJurIVC14}, LOMO \cite{LiaHuZhuLiCVPR15}, HOG3D \cite{KlaMarSchBMCV08} on MARS dataset.}
			\label{table:rank1}
		}
	\end{table}

	\begin{table}[h!]
		\centering
		\begin{tabular}{c|c|c|r}		
			{Feature+Distance}&{Methods}  &{Within } & Across \\ \cline{1-4}
			\cline{1-4}
			\multicolumn{1}{ c  }{\multirow{3}{*}{CNN + Eucl} } &
			\multicolumn{1}{ |c| }{\cite{ZheBieSunWanSuWanTiaECCV16}} & 0.59 & 0.28     \\ \cline{2-4}
			\multicolumn{1}{ c  }{}        &
			\multicolumn{1}{ |c| }{Ours (PairwiseDist)} & 0.59 & 0.29     \\ \cline{2-4}
			\multicolumn{1}{ c  }{}                        &
			\multicolumn{1}{ |c| }{Ours (MembershipS)} & \textbf{0.60} & \textbf{0.29}     \\ \cline{1-4}
			\multicolumn{1}{ c  }{\multirow{3}{*}{CNN + KISSME} } &
			\multicolumn{1}{ |c| }{\cite{ZheBieSunWanSuWanTiaECCV16}} & 0.61 & 0.34  \\ \cline{2-4}
			\multicolumn{1}{ c  }{}                       &
			\multicolumn{1}{ |c| }{Ours (PairwiseDist)} & 0.64 & 0.41   \\ \cline{2-4}
			\multicolumn{1}{ c  }{}                        &
			\multicolumn{1}{ |c| }{Ours (MembershipS)} & \textbf{0.67} & \textbf{0.44} \\
			\cline{1-4}
			\multicolumn{1}{ c  }{\multirow{3}{*}{CNN + XQDA} } &
			\multicolumn{1}{ |c| }{\cite{ZheBieSunWanSuWanTiaECCV16}} & 0.62 & 0.35  \\ \cline{2-4}
			\multicolumn{1}{ c  }{}                      &
			\multicolumn{1}{ |c| }{Ours (PairwiseDist)} & 0.65 & 0.42   \\ \cline{2-4}
			\multicolumn{1}{ c  }{}                        &
			\multicolumn{1}{ |c| }{Ours (MembershipS)} & \textbf{0.68} & \textbf{0.45} \\ \cline{2-4}

			\cline{1-4}
			
		\end{tabular}
		
		
		\caption{The results show performance of our(using pairwise distance and membership score) and state-of-the-art approach \cite{ZheBieSunWanSuWanTiaECCV16} in solving within- and across-camera ReID using average precision on MARS dataset using CNN feature and different distance metrics.
		}
		\label{table:NotionOfMembershipScore}
	\end{table}
	To show how much meaningful the notion of centrality of constrained dominant set is, we  conduct an experiment on the MARS dataset computing the final ranking using the membership score and pairwise distances. The confusion matrix in Fig. \ref{fig:NotionOfMembershipScore1} shows the detail result of both the within cameras (diagonals) and across cameras (off-diagonals), as we consider tracks from each camera as query. Given a query, a set which contains the query is extracted using the constrained dominant set framework.  Note that constraint dominant set comes with the membership scores for all members of the extracted set. We show  in Figure \ref{fig:NotionOfMembershipScore1} the results  based on the final ranking obtained using membership scores (\textbf{left}) and using pairwise Euclidean distance between the query and the extracted nodes(\textbf{right}). As can be seen from the results in Table \ref{table:NotionOfMembershipScore} (average performance) the use of membership score outperforms the pairwise distance approach, since it captures the interrelation among  targets. 
	
	\subsection{Computational Time.} 
\begin{figure}[h!]
		\centering
		\includegraphics[width=1\linewidth,trim=3cm 8cm 4.5cm 9.5cm,clip]{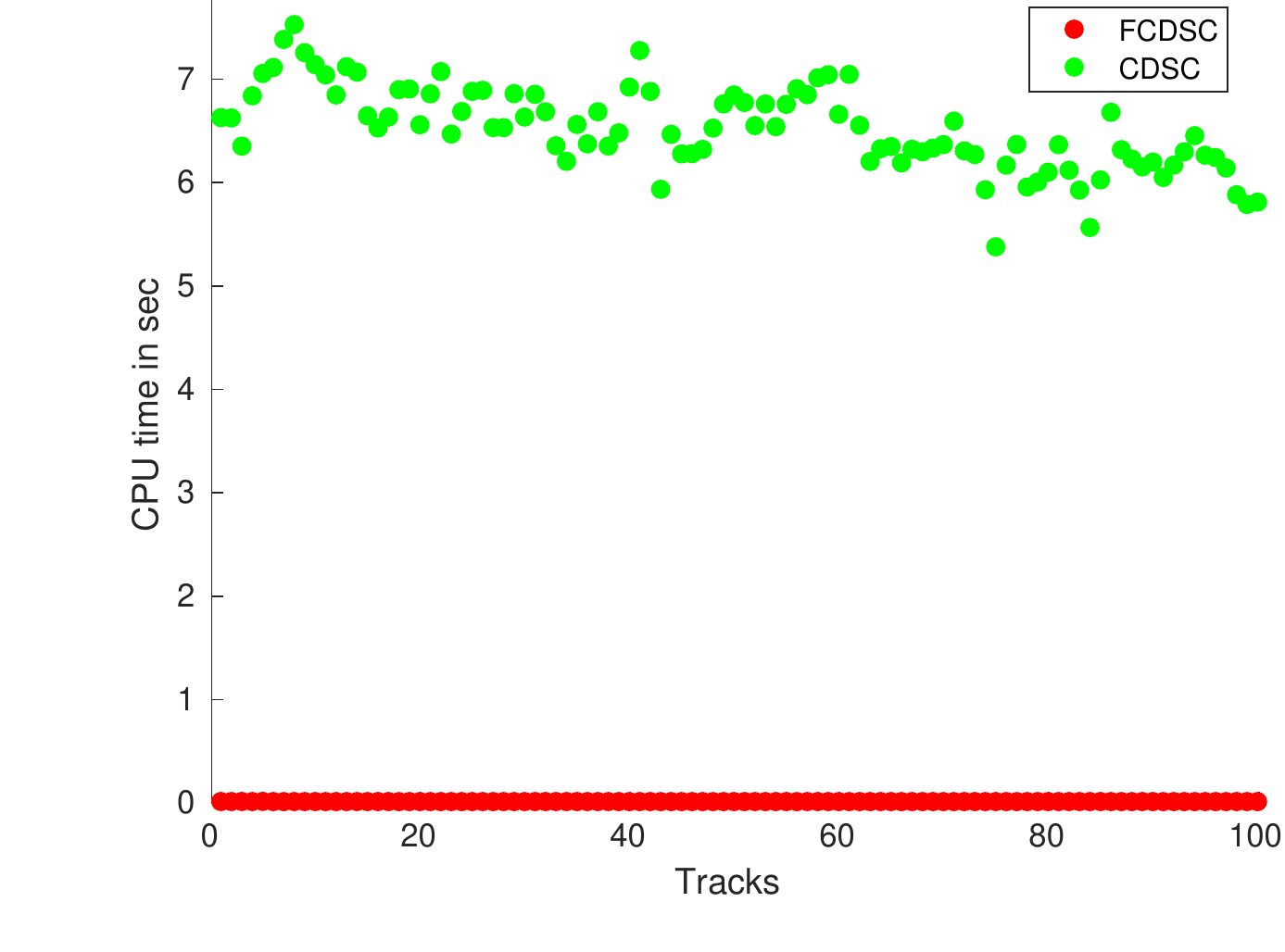}
		\caption{CPU time taken for each track association using our proposed fast approach (FCDSC - fast CDSC) and CDSC.}
		\label{plot:time_plot}
	\end{figure}

\begin{figure}[h!]
		\centering
		\includegraphics[width=1\linewidth,trim=3cm 8cm 4.5cm 9.5cm,clip]{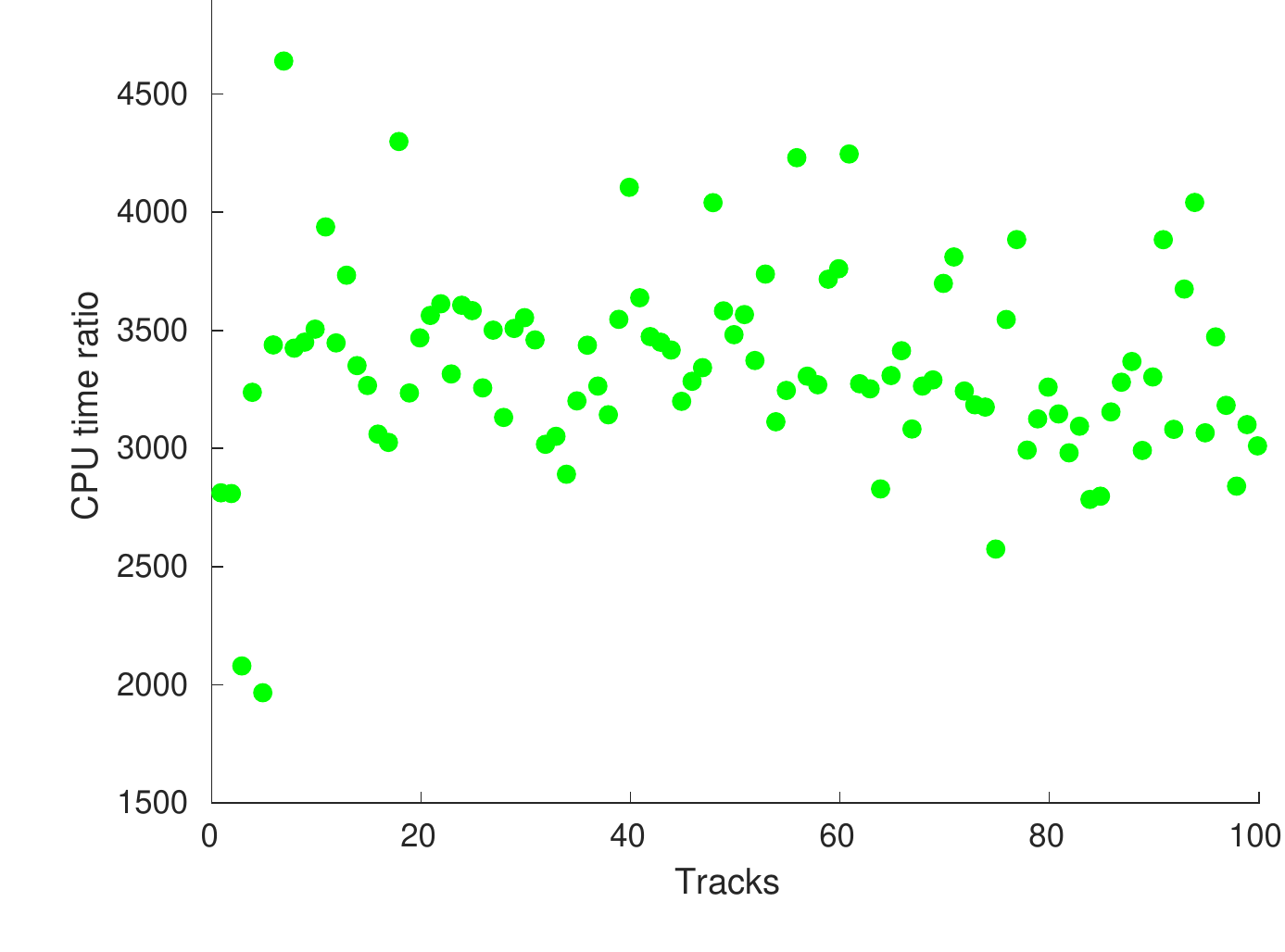}
		\caption{The ratio of CPU time taken between CDSC and proposed fast approach (FCDSC), computed as CPU time for CDSC/CPU time for FCDSC.}
		\label{plot:cpuTimeRatio}
	\end{figure}

\begin{figure*}[h]
		\centering
		\includegraphics[width=1\linewidth]{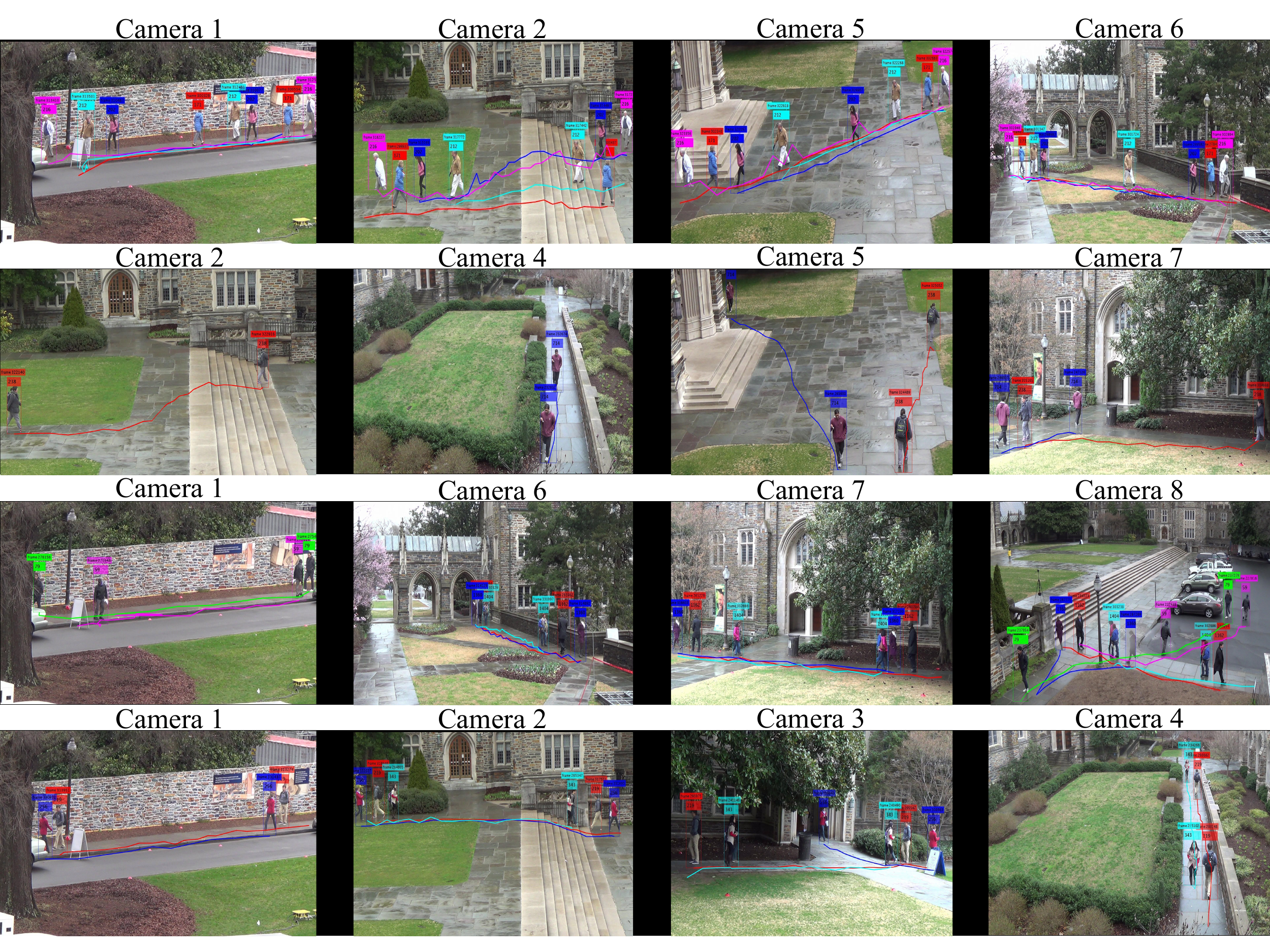}
		\caption{\small Sample qualitative results of the proposed approach on DukeMTMC dataset. Bounding boxes and lines with the same color indicate the same target (Best viewed in color).
		}
		\label{fig:QualitativeResults}
	\end{figure*}

Figure \ref{plot:time_plot} shows the time taken for each track - from 100 randomly selected (query) tracks - to be associated, with the rest of the (gallery) tracks, running CDSC over the whole graph (CDSC without speedup) and running it on a small portion of the graph using the proposed approach (called FCDSC, CDSC with speedup). The vertical axis is the CPU time  in seconds and horizontal axis depicts the track IDs.  As it is evident from the plot,our approach takes a fraction of second  (red points in Fig. \ref{plot:time_plot}). Conversely, the CDSC takes up to 8 seconds for some cases (green points in Fig. \ref{plot:time_plot}). Fig. \ref{plot:cpuTimeRatio} further elaborates how fast our proposed approach is over CDSC, where the vertical axis represents the ratio between CDSC (numerator) and FCDSC (denominator) in terms of CPU time. This ratio ranges from 2000 (the proposed FCDSC 2000x faster than CDSC) to a maximum of above 4500.

	\section{Conclusions} \label{conclusion}
	In this paper we presented a constrained dominant set clustering (CDSC) based framework for  solving multi-target tracking problem in multiple non-overlapping cameras. The proposed method utilizes a three layers hierarchical approach, where within-camera tracking is solved using first two layers of our framework resulting in  tracks for each person, and later in the third layer the proposed across-camera tracker merges tracks of the same person across different cameras. Experiments on a challenging real-world dataset (MOTchallenge DukeMTMCT) validate the effectivness of our model. 
	
	We further perform additional experiments to show effectiveness of the proposed across-camera tracking on one of the largest video-based people re-identification datasets (MARS). Here each query is treated as a constraint set and its corresponding members in the resulting constrained dominant set cluster are considered as possible candidate matches to their corresponding query.
	
	There are few directions we would like to pursue in our future research. In this work, we consider a static cameras with known topology but it is important for the approach to be able to handle challenging scenario, were some views are from cameras with ego motion (e.g., PTZ cameras or taken from mobile devices) with unknown camera topology. Moreover, here we consider features from static images, however, we believe video features which can be extracted using LSTM could boost the performance and help us extend the method to handle challenging scenarios.

    \bibliographystyle{IEEEtran}

	\bibliography{GeoLocDomSetV1}
	
	\newpage

	\vspace*{-2cm}
	\begin{IEEEbiography}[{\includegraphics[width=1.0in,height=1.5in,clip,keepaspectratio]{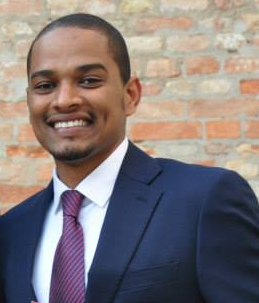}}]{Yonatan Tariku Tesfaye}
		received his BSc degree in computer science from Arba Minch University in 2007. He has worked 5 years at Ethio-telecom as senior programmer and later joined CaFoscari University of Venice and received his MSc degree (with Honor) in computer science in 2014. He is currently a PhD student at IUAV university of Venice starting from 2014. He is now a research assistance, towards his PhD, at Center for Research in Computer Vision at University of Central Florida. His research interests include multi-target tracking, people re-identification, segmentation, image geo-localization, game theoretic model and graph theory.
	\end{IEEEbiography}\vspace*{-1cm}
	\begin{IEEEbiography}[{\includegraphics[width=1.0in,height=1.5in,trim=6cm 3cm 6cm 0.5cm,clip,keepaspectratio]{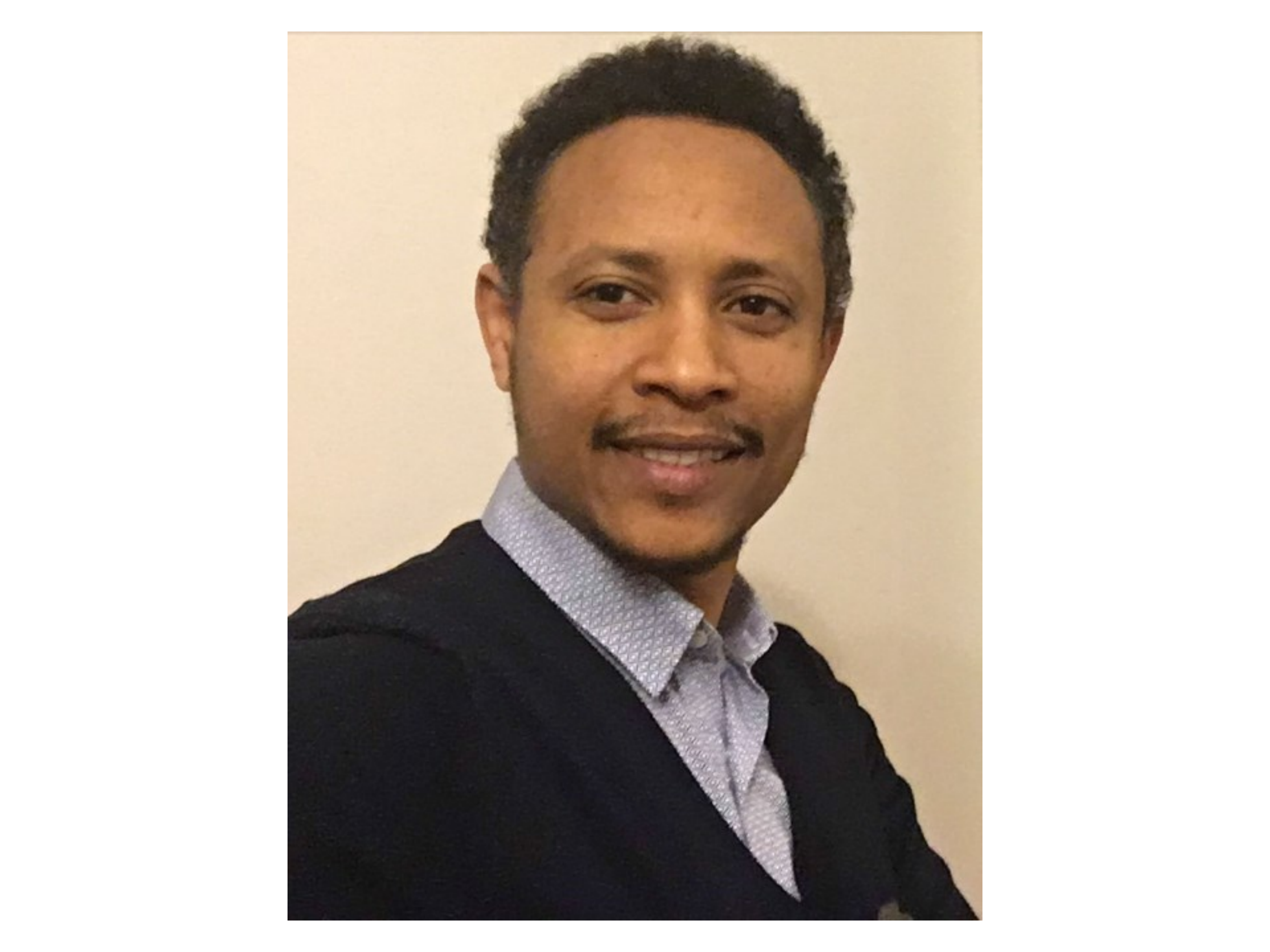}}]{Eyasu Zemene}
		received the BSc degree in Electrical Engineering from Jimma University in 2007, he then worked at Ethio Telecom for 4 years till he joined CaFoscari University (October 2011) where he got his MSc in Computer Science in June 2013. September 2013, he won a 1 year research fellow to work on Adversarial Learning at Pattern Recognition and Application lab of University of Cagliari. Since September 2014 he is a PhD student of CaFoscari University under the supervision of prof. Pelillo. Working towards his Ph.D. he is trying to solve different computer vision and pattern recognition problems using theories and mathematical tools inherited from graph theory, optimization theory and game theory. Currently, Eyasu is working as a research assistant at Center for Research in Computer Vision at University of Central Florida under the supervision of Dr. Mubarak Shah. His research interests are in the areas of Computer Vision, Pattern Recognition, Machine Learning, Graph theory and Game theory.
	\end{IEEEbiography}
	\vspace*{-1cm}
	\begin{IEEEbiography}[{\includegraphics[width=1.0in,height=1.5in,clip,keepaspectratio]{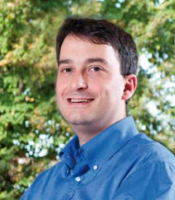}}]{Andrea Prati} Andrea Prati graduated in Computer Engineering at the University of Modena and Reggio Emilia in 1998. He got his PhD in Information Engineering in 2002 from the same University. After some post-doc position at University of Modena and Reggio Emilia, he was appointed as Assistant Professor at the Faculty of Engineering of Reggio Emilia (University of Modena and Reggio Emilia) from 2005 to 2011, and then as Associate Professor at the Department of Design and Planning in Complex Environments of the University IUAV of Venice, Italy. In 2013 he has been promoted to full professorship, waiting for official hiring in the new position. In December 2015 he moved to the Department of Engineering and Architecture of the University of Parma. Author of 7 book chapters, 31 papers in international referred journals (including 9 papers published in IEEE Transactions) and more than 100 papers in proceedings of international conferences and workshops. Andrea Prati is Senior Member of IEEE, Fellow of IAPR ("For contributions to low- and high-level algorithms for video surveillance"), and member of GIRPR.
	\end{IEEEbiography}
	\vspace*{-1cm}
	
	\begin{IEEEbiography}[{\includegraphics[width=1.0in,height=1.5in,clip,keepaspectratio]{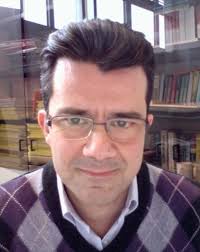}}]{Marcello Pelillo} is Full Professor of Computer Science at CaFoscari University in Venice, Italy, where he directs the European Centre for Living Technology (ECLT) and leads the Computer Vision and Pattern Recognition group. He held visiting research positions at Yale University, McGill University,the University of Vienna, York University (UK), the University College London, and the National ICT Australia (NICTA). He has published more than 200 technical papers in refereed journals, handbooks, and conference proceedings in the areas of pattern recognition, computer vision and machine learning. He is General Chair for ICCV 2017 and has served as Program Chair for several conferences and workshops (EMMCVPR, SIMBAD, S+SSPR, etc.). He serves (has served) on the Editorial Boards of the journals IEEE Transactions on Pattern Analysis and Machine Intelligence (PAMI), Pattern Recognition, IET Computer Vision, Frontiers in Computer Image Analysis, Brain Informatics, and serves on the Advisory Board of the International Journal of Machine Learning and Cybernetics. Prof. Pelillo is a Fellow of the IEEE and a Fellow of the IAPR.
	\end{IEEEbiography} 
	\vspace*{-1cm}
	\begin{IEEEbiography}[{\includegraphics[width=1.0in,height=3in,trim=0cm 0cm 0cm 0cm, clip,keepaspectratio]{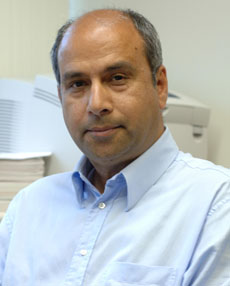}}]{Mubarak Shah,} the Trustee chair professor of computer science, is the founding director of the Center for Research in Computer Vision at the University of Central Florida (UCF). He is an editor of an international book series on video computing, editor-in-chief of Machine Vision and Applications journal, and an associate editor of ACM Computing Surveys journal. He was the program cochair of the IEEE Conference on Computer Vision and Pattern Recognition (CVPR) in 2008, an associate editor of the IEEE Transactions on Pattern Analysis and Machine Intelligence, and a guest editor of the special issue of the International Journal of Computer Vision on Video Computing. His research interests include video surveillance, visual tracking, human activity recognition, visual analysis of crowded scenes, video registration, UAV video analysis, and so on. He is an ACM distinguished speaker. He was an IEEE distinguished visitor speaker for 1997-2000 and received the IEEE Outstanding Engineering Educator Award in 1997. In 2006, he was awarded a Pegasus Professor Award, the highest award at UCF. He received the Harris Corporations Engineering Achievement Award in 1999, TOKTEN awards from UNDP in 1995, 1997, and 2000, Teaching Incentive Program Award in 1995 and 2003, Research Incentive Award in 2003 and 2009, Millionaires Club Awards in 2005 and 2006, University Distinguished Researcher Award in 2007, Honorable mention for the ICCV 2005 Where Am I? Challenge Problem, and was nominated for the Best Paper Award at the ACM Multimedia Conference in 2005. He is a fellow of the IEEE, AAAS, IAPR, and SPIE.
	\end{IEEEbiography}
	
	\newpage
	
	\section*{Appendix}
	
	\vspace*{1cm}
	
	\begin{proposition}
		Given an affinity A and a distribution $\x \in \Delta$, if $(A\x)_i             > \x'A\x - \alpha \x'_{\mathcal{Q}} \x_{\mathcal{Q}}, \mbox{for} i \notin \sigma(\x)$,
		
		\begin{enumerate}
			\item $\x$ is not the maximizer of the parametrized quadratic program of (\ref{eqn:parQP})
			\item $e_i$ is a \textbf{dominant distribution} for $\x$
		\end{enumerate}
		
	\end{proposition}
	
	\vspace*{1cm}
	
	\proof To show the first condition holds: Let's assume $\x$ is a KKT point
	
	\[
	\x\T ({\mat{A}} - \alpha I_{\mathcal{Q}}) \x = \sum\limits_{i=1}^{n} x_i [(A - \alpha I_{\mathcal{Q}}) \x]_i\]
	Since x is a KKT point
	\[ \x\T ({\mat{A}} - \alpha I_{\mathcal{Q}}) \x = \sum\limits_{i=1}^{n} x_i * \lambda/2 = \lambda/2\]
	From the second condition, we have: 
	\[ [(A - \alpha I_{\mathcal{Q}}) \x]_i \le \lambda/2 = \x\T ({\mat{A}} - \alpha I_{\mathcal{Q}}) \x \]
	Since $i \notin \sigma(\x)$
	\[ (A\x)_i \le  \x\T ({\mat{A}} - \alpha I_{\mathcal{Q}}) \x \]
	Which concludes the proof showing that the inequality does not hold.
	
	For the second condition, if $e_i$ is a \textbf{dominant distribution} for $\x$, it should satisfy the inequality
	
	\[ \left\{ e_i\T ({\mat{A}} - \alpha I_{\mathcal{Q}}) \x \right\} > \left\{\x\T ({\mat{A}} - \alpha I_{\mathcal{Q}}) \x \right\}\]
	Since $i \notin \sigma(\x)$
	
	\[ (Ax)_i > \left\{\x\T ({\mat{A}} - \alpha I_{\mathcal{Q}}) \x \right\}\]
	Which concludes the proof
	
	\endproof

\end{document}